\documentclass[preprint,12pt]{elsarticle}

\usepackage{amssymb}
\usepackage{amsmath}

\usepackage{algorithm}
\usepackage{algpseudocode}

\usepackage[table]{xcolor}
\usepackage{collcell}
\usepackage{array}
\usepackage{multirow}
\usepackage{rotating}

\usepackage{array}

\usepackage{caption}
\usepackage{graphicx}
\usepackage{calc} 

\usepackage{hyperref}
\usepackage{booktabs}

\newif\ifshowchanges\showchangestrue
\definecolor{changedblue}{RGB}{0,70,200}
\definecolor{deletedred}{RGB}{200,0,0}
\usepackage[normalem]{ulem} 

\newcommand{\delsout}{\bgroup\ULdepth=-0.8ex\ULset}

\ifshowchanges\usepackage[colorinlistoftodos,textsize=scriptsize,color=orange!25]{todonotes}
\else\usepackage[disable]{todonotes}\fi
\definecolor{heatmap0}{RGB}{100,140,190} 
\definecolor{heatmap1}{RGB}{120,160,205} 
\definecolor{heatmap2}{RGB}{140,180,220} 
\definecolor{heatmap3}{RGB}{160,200,235} 
\definecolor{heatmap4}{RGB}{180,210,240} 
\definecolor{heatmap5}{RGB}{220,220,225} 
\definecolor{heatmap6}{RGB}{235,190,185} 
\definecolor{heatmap7}{RGB}{220,150,140} 
\definecolor{heatmap8}{RGB}{200,110,100} 
\definecolor{heatmap9}{RGB}{190,80,70}   

\definecolor{oursgreen}{RGB}{220, 235, 220}
\newcommand{\best}[1]{{\bfseries\boldmath #1}}

\graphicspath{ {imgs/} }

\journal{Neurocomputing}

\begin{document}

\begin{frontmatter}



\title{In Two Minds about Lifelong Learning: Exploring Hemispheric Redundancy and Specialisation in Neural Models}

\author[1]{\texorpdfstring{Benjamin Smith\corref{cor1}}{Benjamin Smith}}
\ead{benjamin.smith@monash.edu}
\author[1]{Levin Kuhlmann}
\ead{levin.kuhlmann@monash.edu}
\author[3]{Kaushik Roy}
\ead{kaushik.roy@csiro.au}
\author[1
,2]{Gideon Kowadlo}
\ead{gideon@cerenaut.ai}
\cortext[cor1]{Corresponding author}

\affiliation[1]{organization={Data Science and AI, Monash University},
            addressline={Wellington Road}, 
            city={Clayton},
            postcode={3800}, 
            state={Victoria},
            country={Australia}}
\affiliation[2]{organization={Cerenaut},
            state={Victoria},
            country={Australia}}
\affiliation[3]{organization={CSIRO Robotics, CSIRO}, 
country={Australia}}


\begin{abstract}
Persistent intelligent systems require the ability to learn continually, but current machine learning approaches face significant challenges in this area compared to biological learning systems. Machine learning algorithms typically trade off retention of previously learned information and adaptation to new or changing data patterns. When continual learning capabilities are absent, algorithms must undergo retraining using the entire data set, an approach that becomes impractical when original training data are unavailable due to storage constraints, financial or computational costs, or privacy restrictions. However, biological animals can learn continually, without experiencing catastrophic forgetting. This paper attempts to build a high-level framework for how animals learn and preserve knowledge by modelling neural components and states that are known to be related to memory consolidation. We focus on three concepts: experience replay, REM sleep, and bilaterality. We propose 4MAS (4 Module Awake/Sleep), a novel macroarchitecture demonstrating how machine learning models might benefit from asymmetric hemispheres, each with their own long- and short-term memory mechanisms, and how a period of sleep between incremental learning tasks might benefit memory consolidation. Finally, we present results showing that our architecture achieves statistical improvements over established generative replay baselines on Split-MNIST (98.3\%) and Split-Fashion-MNIST (84.9\%), while reducing forgetting by more than half relative to the strongest replay baseline and demonstrating monotonic capacity scaling on Split-CIFAR-100 (29.29\%).
\end{abstract}

\begin{graphicalabstract}
\includegraphics[width=1\linewidth]{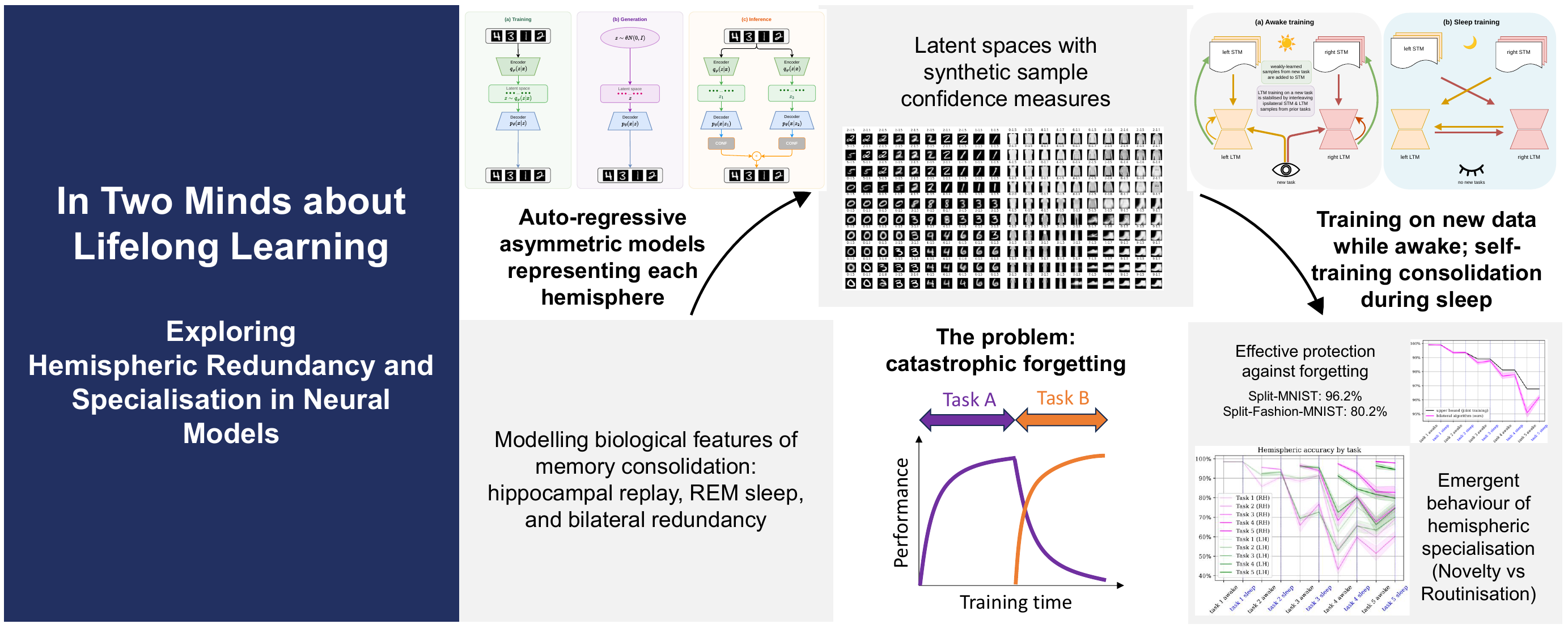}
\end{graphicalabstract}



\end{frontmatter}



\section{Introduction}
\label{sec:introduction}

Contemporary artificial neural networks suffer from catastrophic forgetting (or catastrophic interference) \cite{wickramasinghe_continual_2024,mccloskey_catastrophic_1989}, where learning new tasks rapidly erodes previously acquired skills. While standard training assumes independent, identically distributed (i.i.d.) data, real-world applications present sequentially correlated distributions where class presentation frequencies vary arbitrarily.

This forgetting stems from the stability-plasticity dilemma: adapting weights to new data modifies parameters that are critical for earlier tasks \cite{kemker_measuring_2017,kirkpatrick_overcoming_2017}. Joint training across all data streams is fundamentally impossible in continual learning settings where future data are unavailable and past data are restricted by storage or privacy constraints. Naive sequential fine-tuning, while feasible, suffers from severe catastrophic forgetting. Continual learning algorithms seek to resolve this dilemma without full dataset retraining.

In contrast, biological brains mitigate forgetting through specialised neural architectures and consolidation states. The mammalian hippocampus uses experience replay to weave together recent and distant memories, transferring knowledge to the neocortex during sleep \cite{corkin_whats_2002,giri_hippocampal_2019,hayes_replay_2021}. Furthermore, sleep phases like rapid-eye-movement (REM) sleep exhibit highly coordinated bilateral (inter-hemispheric) activity, which contrasts with the unihemispheric states of slow-wave sleep \cite{tamaki_night_2016,ghosh_running_2022}. Finally, there is evidence that biological brains leverage two lateralised cortical hemispheres that acquire distinct representations, with the right hemisphere handling novel information and the left hemisphere optimised for stable routine tasks \cite{koziol_noveltyroutinization_2014,prat_getting_2023}.

Existing generative replay methods such as Generative Replay struggle to scale because a single generator must capture all historical distributions, leading to compounding representational drift \cite{vandeven_braininspired_2020}. To address these limitations, we propose 4MAS (4 Module Awake/Sleep), a novel continual learning macroarchitecture that implements two asymmetric hemispheres—each with dedicated short- and long-term memory (STM and LTM) modules—and an explicit offline sleep phase for cross-hemispheric consolidation; coarsely representing hippocampus and neocortex, respectively. By distributing replay across two specialised generators and anchoring the latent space during sleep, 4MAS minimises representational drift.

Our main contributions are:
\begin{itemize}
    \item We propose a dual-hemisphere continual learning architecture (4MAS) that splits generative replay across specialised exploratory and conservative models, mimicking biological lateralisation.
    \item We introduce an explicit wake-sleep training cycle that uses cross-hemispheric consolidation (mutual fine-tuning during a simulated sleep phase) to stabilise latent spaces and prevent representational drift.
    \item We demonstrate that 4MAS achieves competitive Class-IL accuracies on Split-MNIST ($98.3\%$), Split-Fashion-MNIST ($84.9\%$), and Split-CIFAR-100 ($29.29\%$).
\end{itemize}

\subsection{Biological motivation}\label{sec:biomotivation}
Inspired by this neurobiological template, 4MAS employs two generative long-term memories as an asymmetrical ensemble: one retains plasticity for novel task acquisition, while the other stabilises historical patterns. Generative memories are paired with short-term memory buffers, allowing for rehearsal akin to hippocampal replay \cite{olafsdottir_role_2018}. Training follows an ultradian-inspired cycle, alternating between a unihemispheric awake phase for separate task learning in both hemispheres and a bilateral sleep phase for cross-hemispheric memory harmonisation.

\section{Background and related work}\label{sec:background}

Continual learning is typically evaluated under three scenarios \cite{ven_three_2019} of increasing difficulty:
\begin{itemize}
\item \textbf{Task-IL (Task Incremental Learning):} The model learns a sequence of tasks with distinct data distributions and separate output spaces. The task identity is explicitly provided during both training and inference, allowing the model to select task-specific parameters or heads.
\item \textbf{Domain-IL (Domain Incremental Learning):} The model is exposed to a sequence of tasks with varying input distributions that share a common output space. The task identity is never provided, requiring the model to adapt to changing domains without contextual routing.
\item \textbf{Class-IL (Class Incremental Learning):} The most challenging paradigm, where both input distributions and output spaces vary across tasks. No task identifiers are available at training or inference. The model must classify inputs across all classes seen so far, making it highly susceptible to catastrophic forgetting.
\end{itemize}

When evaluating these models, performance is typically benchmarked against fine-tuning (sequential training without forgetting mitigation; lower bound) and joint-training (simultaneous training on all data; upper bound). To bridge the gap to joint-training under Class-IL constraints, algorithms are commonly grouped into regularisation, parameter isolation, and rehearsal.

\subsection{Regularisation}
Regularisation methods constrain learning to preserve prior knowledge, either by protecting parameters critical to earlier tasks (parameter regularisation) or by constraining the network's input-output function (functional regularisation) \cite{vandeven_three_2022}. Early approaches froze lower layers of a network after training on a task \cite{gutstein_knowledge_2008}. Modern algorithms estimate parameter importance by calculating contribution to loss reduction (e.g., Synaptic Intelligence \cite{zenke_continual_2017}), approximating Bayesian inference (e.g., Elastic Weight Consolidation \cite{kirkpatrick_overcoming_2017}), or evaluating output function sensitivity (e.g., Memory Aware Synapses \cite{aljundi_memory_2018}). Recent methods perform geometric analysis to locate stable flat minima in the distribution manifold \cite{mirzadeh_understanding_2020,shi_overcoming_2021} or approximate prior losses using the Hessian matrix eigenvalues \cite{kong_overcoming_2024}. Functional regularisation instead distils the responses of the previous model on current data, as in Learning without Forgetting \cite{li_learning_2018}.

While regularisation is highly effective in Task-IL \cite{lange_continual_2021}, it struggles in Class-IL \cite{vandeven_three_2022}. The strict constraints designed to protect existing knowledge prevent the model from adapting to novel classes, leading to representational paralysis as constraints accumulate over sequential tasks \cite{shi_overcoming_2021}.

\subsection{Parameter isolation}
Parameter isolation limits training to a subset of parameters or dynamically expands the network structure. A common design adds task-specific output heads to a pretrained feature-extractor backbone \cite{razavian_cnn_2014,donahue_decaf_2013}. Under network expansion, the network freezes historical weights and adds new neurons or columns for each task \cite{terekhov_knowledge_2015,rusu_progressive_2022}. Another branch of research leverages Adaptive Resonance Theory to dynamically create category nodes as new distributions appear \cite{grossberg_adaptive_2013}.

However, these approaches struggle in Class-IL settings because they require task identity during inference to route data to the appropriate sub-network. Without explicit task identifiers, the model lacks an intrinsic mechanism for choosing the correct task pathway. Although sparse neural activation in large networks can mitigate routing issues \cite{shazeer_outrageously_2017}, parameter isolation remains difficult to scale without introducing capacity exhaustion or inference-routing failures \cite{chen_learning_2025,omi_load_2025}.

\subsection{Rehearsal: replay and generative replay}
Rehearsal techniques interleave historical data with new task inputs. Experience replay stores a memory buffer of real samples from earlier tasks, and Gradient Episodic Memory (GEM) uses those stored examples as inequality constraints on each update so the loss on previous tasks does not increase, which reduces forgetting while still allowing positive backward transfer
\cite{lopez-paz_gradient_2017}. Maximally Interfered Retrieval (MIR) instead keeps a finite replay memory, estimates the parameter update from the current batch, and then replays the buffered samples whose losses would increase the most under that update, making it distinct from GEM because it prioritises most-interfered samples rather than enforcing explicit gradient constraints \cite{aljundi_online_2019}. AdaER further adapts replay by using Contextually-Cued Memory Recall to select memories based on both data-conflicting and task-conflicting cues, and it also updates the buffer with Entropy-Balanced Reservoir Sampling to keep a more balanced, informative memory, distinguishing it from MIR’s interference-only retrieval and GEM’s constraint-based updates \cite{li_adaer_2024}.

While buffer-based replay is effective, it scales poorly because representing complex, high-dimensional distributions requires a prohibitive number of stored samples \cite{balaji_effectiveness_2020}.

Generative replay avoids storing real data by training a generative model to synthesise historical inputs \cite{shin_continual_2017}. To scale beyond simple datasets, models reconstruct latent feature representations rather than raw training samples. For example, Generative Feature Replay (GFR) uses a feature extractor to train a generator on latent distributions \cite{liu_generative_2020}, mimicking biological systems where memory replay occurs at representational rather than raw sensory levels \cite{rolls_ventromedial_2024}. Brain-Inspired Replay (B-IR) combines generative replay with parameter isolation to achieve state-of-the-art results \cite{vandeven_braininspired_2020}. Nonetheless, generative models suffer from representational drift as the feature extractor updates over time, requiring distillation constraints or frozen features to anchor the latent space \cite{khan_looking_2024}.

\section{Methods}\label{sec:methods}

\begin{figure}
  \centering
  \includegraphics[width=1\linewidth]{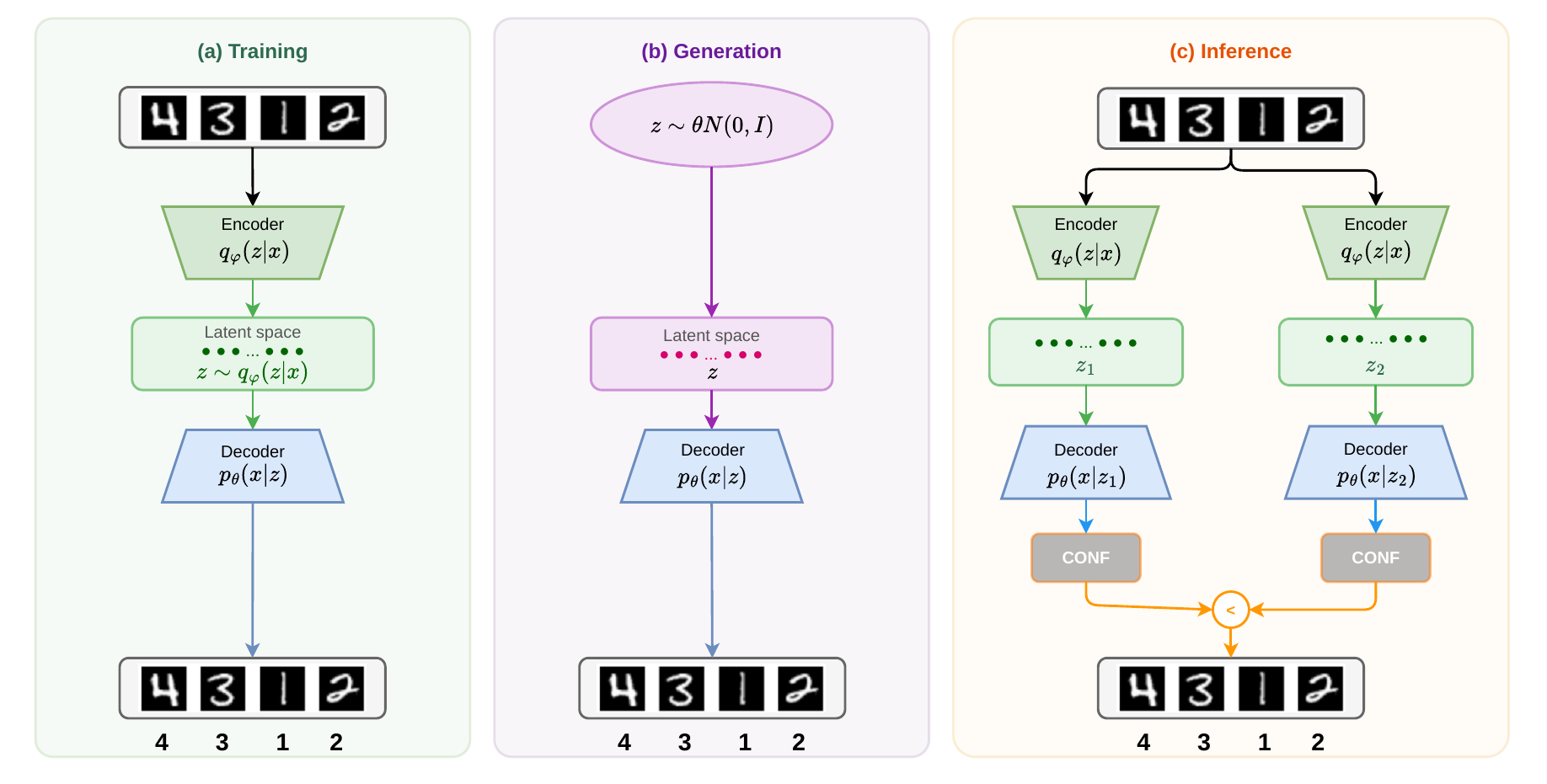}
  \caption{\label{fig:architecture}LTM inputs and outputs: (a) LTM training on image-label pairs for each hemisphere during awake and sleep phases. (b) Sample generation with temperature $\tau$ from each hemisphere provides two synthetic input sources for subsequent LTM training tasks. (c) During inference, both hemispheres process inputs and the highest confidence classification is selected.}
\end{figure}

We present 4MAS (4 Module Awake/Sleep), a macroarchitecture for continual learning designed to model biological memory consolidation processes. The system consists of two lateralised hemispheres. Each hemisphere contains:
\begin{enumerate}
    \item A generative Long-Term Memory (LTM) representing the neocortex, which learns task distributions and classifies incoming data. Figure~\ref{fig:architecture} shows how the LTM is connected throughout training and inference.
    \item A Short-Term Memory (STM) buffer representing the hippocampus, which stores a small set of episodic exemplars. Figure~\ref{fig:training_cycles} shows training pathways between STMs and LTMs.
\end{enumerate}

The architecture restricts data flow to a biologically inspired model where task acquisition is unihemispheric (Figure~\ref{fig:training_cycles}a) and offline consolidation is bilateral (Figure~\ref{fig:training_cycles}b), facilitating knowledge transfer and specialisation.

\subsection{Benchmarks and experimental design}\label{sec:benchmarks}
\begin{figure}
\centering
\includegraphics[width=1\linewidth]{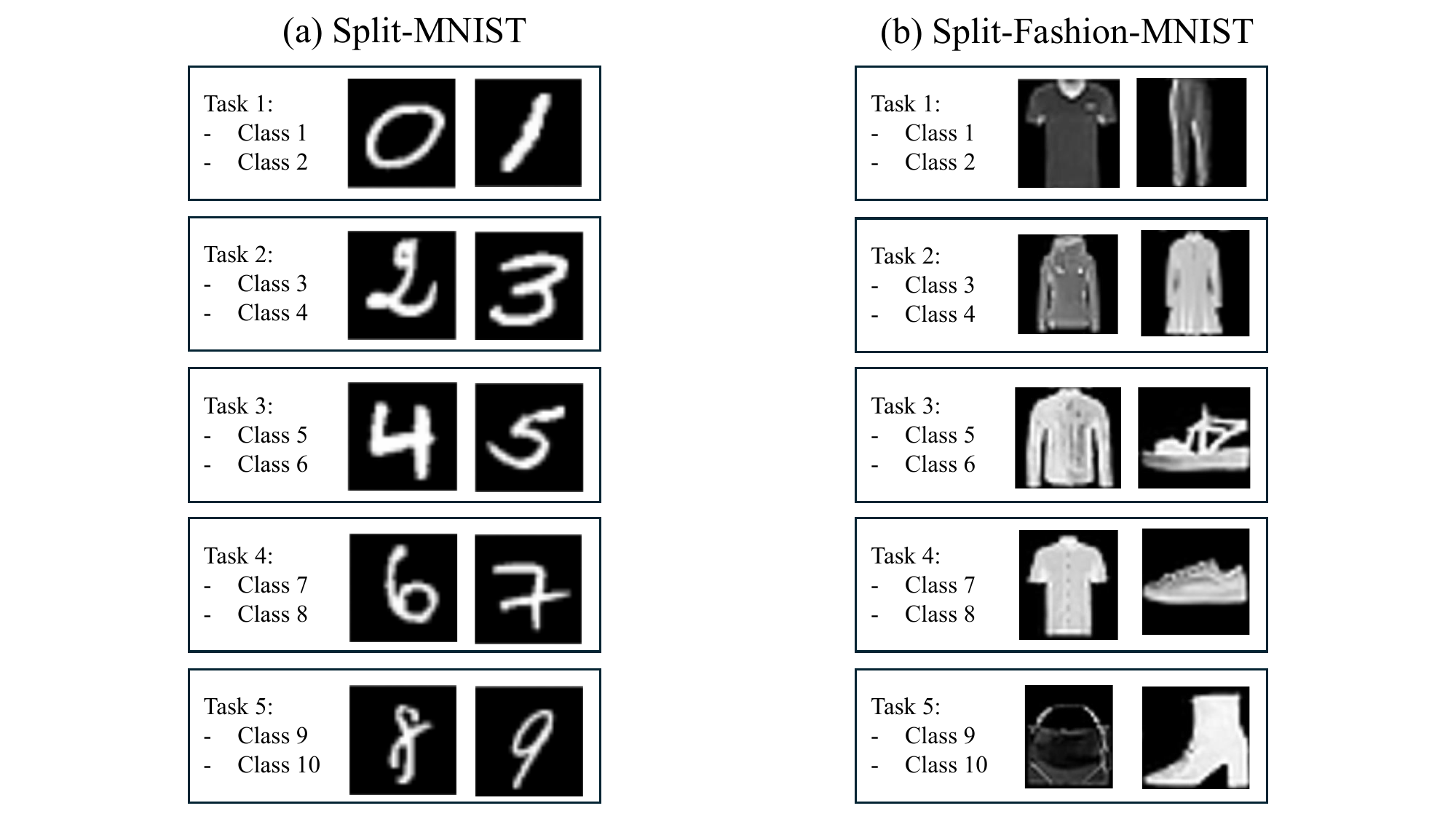}
\caption{\label{fig:datasets}MNIST datasets are split into 5 tasks of 2 classes each for the Class-IL scenario: (a) Split-MNIST, (b) Split-Fashion-MNIST.}
\end{figure}

To evaluate 4MAS under the Class-IL constraint, we employ Split-MNIST \cite{lecun_gradientbased_1998,vandeven_three_2022} and Split-Fashion-MNIST \cite{xiao_fashionmnist_2017,sokar_spacenet_2021}, shown in Figure~\ref{fig:datasets}. The standard 10-class datasets are split into 5 sequential tasks of 2 classes each, presented without task identifiers at both training and inference. Hyperparameter tuning and initial method development were conducted primarily on these two MNIST variants. To test scalability and out-of-the-box generalizability, we then expanded evaluation to the more challenging Split-CIFAR-100 dataset \cite{vandeven_three_2022}. Whereas Split-MNIST and Split-Fashion-MNIST are trained end-to-end from raw pixels using task-agnostic autoencoders, CIFAR-100 requires visual feature abstraction. Following the established continual learning protocol of Van de Ven et al.~\cite{vandeven_braininspired_2020}, we employ a residual convolutional encoder (ResNet backbone). The early convolutional layers are trained during Task 1 (classes 0–9) and subsequently frozen for all subsequent tasks ($i \ge 2$) to provide a stable representation space. The bilateral VAE generators, STM buffers, and classification heads operate directly on these extracted convolutional features. 

Unless otherwise stated, each configuration is run for 10 random seeds, and we report the mean and standard error of the mean of each metric across seeds. The model trains for 10 epochs per class, with a synthetic epoch length of 10,000 samples. Ablation studies disentangle the contributions of the STM buffers, the dual-hemisphere ensemble, and the sleep phase. Hyperparameter search details can be reviewed in \ref{sec:appendix_hyperparameters}.

We swept the total trainable parameters for each method to find the best-performing capacity, stated alongside every entry in Table~\ref{tab:dataset_performance}. The peak capacity varies by method: EWC, oEWC, SI and LwF showed no measurable change in accuracy across the capacities we swept, their failure under Class-IL being structural rather than a shortage of capacity, so the smallest configuration is reported for each. On Split-CIFAR-100 the best 4MAS and B-IR configurations both fall in the largest tier tested, giving a comparison at $\sim$140M; this label denotes a tier spanning 130.8M to 147.8M parameters, within which 4MAS (130.8M) is smaller than B-IR (143.8M). Capacity sweeps are reported in \ref{subsec:app_cifar_scaling}. All baselines are run using the reference implementations of \cite{vandeven_three_2022}, following their published protocol, including convolutional layers pre-trained on CIFAR-10 and frozen during incremental training on Split-CIFAR-100, with hyperparameters tuned for best performance.

\subsection{Sleep/wake phases}
\begin{figure}
\centering
\includegraphics[width=1\linewidth]{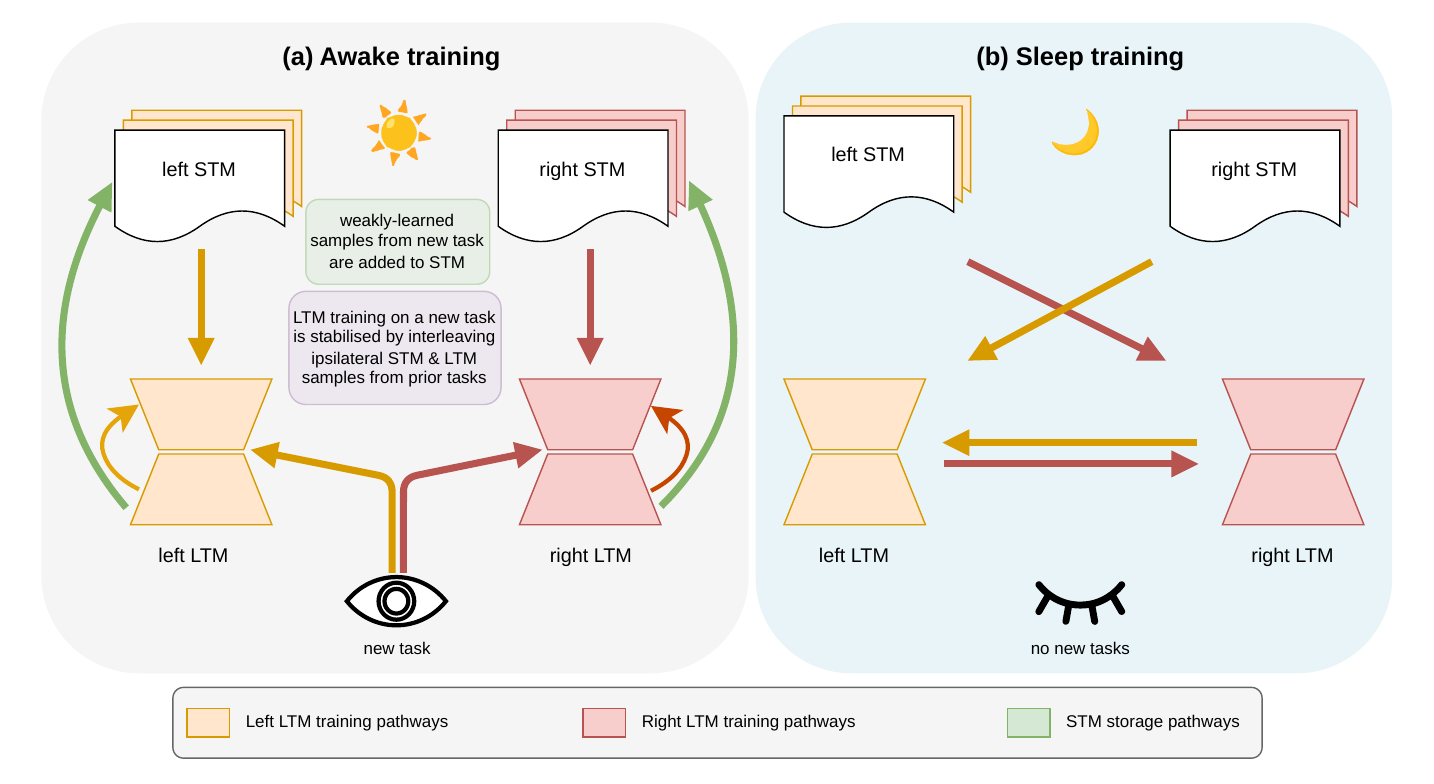}
\caption{\label{fig:training_cycles}Awake/sleep training phases: (a) Awake training: LTMs learn new task data interleaved with STM samples. (b) Sleep training: LTMs are fine-tuned on contralateral representations.}
\end{figure}

The training cycle alternates between a unihemispheric ``awake'' phase and a bilateral ``sleep'' phase for each task, Figure~\ref{fig:training_cycles}.
During the awake phase detailed in Algorithm~\ref{alg:awake}, each LTM hemisphere is trained on three randomly interleaved sources: new task data, exemplars stored in the ipsilateral STM from prior tasks, and self-generated synthetic samples. After training, the LTM evaluates the training data and updates its STM buffer with samples that meet the selection criteria (Section~\ref{sec:stm}).

During the sleep phase detailed in Algorithm~\ref{alg:sleep}, LTMs undergo fine-tuning on contralateral representations (generations from the opposite LTM and exemplars from the opposite STM) during a period where no new training data are available. To allow subtle adjustments to internal representations without destroying learned task-specific parameters, the sleep learning rate is scaled down by a multiplier $\lambda=0.1$, mimicking the lower firing rates observed across brain regions during REM sleep \cite{niethard_sleepstagespecific_2016}.

\begin{algorithm}[H]
    \caption{Incremental task -- awake phase}
    \label{alg:awake}
    \begin{algorithmic}[1]
    \State \textbf{Input:} Task $t\in [1,\infty)$, learning rate $\alpha$, model parameters $\theta_{left}$, $\theta_{right}$, memories $\mathcal{M}_{left}$, $\mathcal{M}_{right}$, classes seen $C$, memory capacity $M$.
    \For{$h \in \{left, right\}$}
        \State $w \leftarrow \{w_i\}_{i=1}^C$ where $w_i = 1/C$
        \For{$e = 1 \dots E_{awake}$}
            \State Draw batch $\beta_t$ from $\mathcal{D}_t$.
            \State \textbf{if} $t > 1$ \textbf{then} generate replay $\mathcal{G}_h \sim p_{\theta_h}(x|z)$ and set $\beta \leftarrow \beta_t \cup \mathcal{G}_h \cup \mathcal{M}_h$ \textbf{else} $\beta \leftarrow \beta_t$
            \State $\beta \leftarrow \mathrm{class\_weighted\_rebalance}(\beta, w)$
            \State $\theta_h \leftarrow \theta_h - \alpha \nabla_\theta \mathcal{L}(\theta_h; \beta)$
            \State $g \leftarrow \mathrm{bincount}\big(\arg\max_c \hat{y}(\mathcal{G}_h)\big)$
            \State $w \leftarrow \{1 / (g_i + \varepsilon)\}_{i=1}^C$ \label{line:inv_bincount}
        \EndFor
        \For{$i = 1 \dots C$}
            \State $\mathcal{P}_{h, i} \leftarrow \mathrm{Oversample}(\mathcal{D}_{t, i} \cup \mathcal{M}_{h, i})$ \Comment{Expand candidate pool for class $i$}
            \State $\mathcal{M}_{h, i} \leftarrow \mathrm{TopK}_{\lfloor M/C \rfloor}\Big(\mathcal{P}_{h, i}, \, CONF(\hat{y})\Big)$ \Comment{Select top-k highest confidence}
        \EndFor
        \State $\mathcal{M}_h \leftarrow \bigcup_{i=1}^C \mathcal{M}_{h, i}$
    \EndFor
    \end{algorithmic}
\end{algorithm}

\begin{algorithm}
    \caption{Incremental task -- sleep phase}
    \label{alg:sleep}
    \begin{algorithmic}[1]
    \State \textbf{Input:} Task $t\in [1,\infty)$, learning rate $\alpha$, models $\theta_{left}$, $\theta_{right}$, memories $\mathcal{M}_{left}$, $\mathcal{M}_{right}$, sleep learning multiplier $\lambda=0.1$.
    \For{$h$ in $\{left, right\}$}
        \State $w \leftarrow \{w_i\}_{i=1}^C$ where $w_i=1/C$
        \For{$e = 1:E_{sleep}$}
            \State $\beta \leftarrow \theta_h(z) \cup \mathcal{M}_h$
            \State $\tau_h \leftarrow \theta_h - \lambda\alpha\nabla_\tau \mathcal{L}(\theta_h; \beta)$.
        \EndFor
    \EndFor
    \end{algorithmic}
\end{algorithm}

\subsection{Long-term memories}
Each LTM is both a generator and a classifier within a single variational auto-encoder (VAE). The classification vector is appended to the input image, and the model is trained to reconstruct the joint vector $x \frown y$. This joint parameter space ensures that synthetic images and labels are tightly coupled. Let $\hat{y}$ denote the reconstructed label channel, normalised over the $C$ classes seen so far. We define prediction confidence as the negative cross-entropy of $\hat{y}$ against its own arg-max class:
\begin{equation}
    CONF(\hat{y}) = -\mathcal{L}_{CE}\big(\hat{y}, \mathrm{onehot}(\arg\max_c \hat{y}_c)\big)
                  = \log \max_c \hat{y}_c
    \label{eq:confidence}
\end{equation}
$CONF \in (-\infty, 0]$, with larger values indicating higher confidence. Since
$\mathcal{L}_{CE}$ is the same term the VAE minimises on the label channel, Equation~\ref{eq:confidence} doubles as a measure of reconstruction quality and requires no separate discriminative head, providing stability when learning new tasks \cite{skiers_joint_2025}. Confidence is used in
three places: rejecting low-confidence generations during replay, ranking candidates for STM storage (Section~\ref{sec:stm}), and arbitrating between hemispheres at inference.

\begin{figure}
\centering
    \begin{minipage}[b]{0.47\textwidth}
    \centering
    \includegraphics[width=\linewidth]{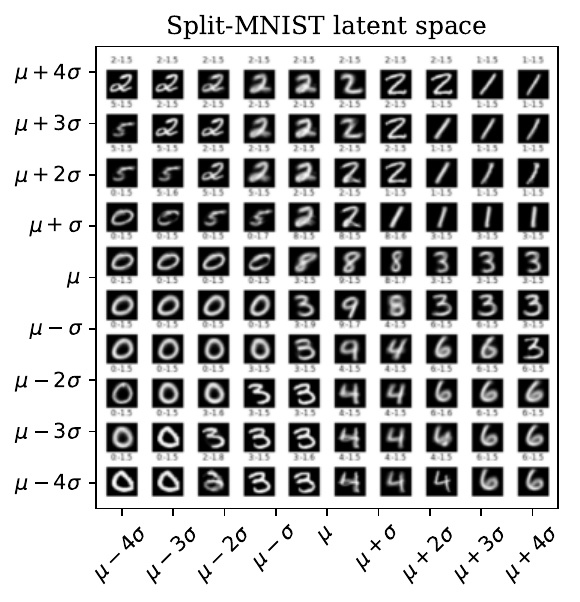}
    \end{minipage}
    \hfill
    \begin{minipage}[b]{0.47\textwidth}
    \centering
    \includegraphics[width=\linewidth]{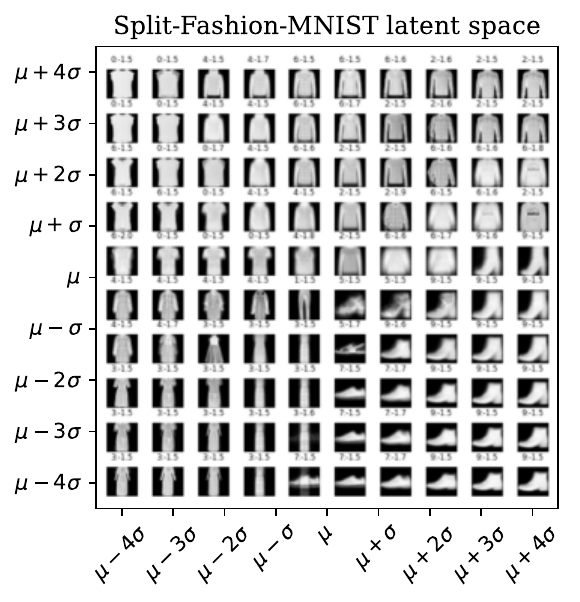}
    \end{minipage}  
  \caption{\label{fig:generatedmnist}2-dimensional mapping of the LTM's latent space after training. $\tau$ scales the sampling radius in units of the prior standard deviation. High temperature $\tau>1$ increases the diversity of output representations (16 latent dimensions used in final configurations).}
\end{figure}

Standard VAE latent sampling $z \sim \mathcal{N}(0,I)$ can suffer from posterior collapse, producing low-variance synthetic samples that degrade replay quality over sequential tasks. To resolve this, we apply a temperature gain $\tau$ to the latent coordinates during sampling: $z = \tau X, X \sim \mathcal{N}(0,I)$ \cite{zhou_parallel_2024}. Setting $\tau > 1$ expands the explored region of the latent space, generating sharper and more diverse samples, Figure~\ref{fig:generatedmnist}. During replay, low-confidence generations near class boundaries are rejected to avoid interpolating between classes. At inference, inputs are processed by both LTMs, and the class prediction from the hemisphere with higher confidence is selected. To maintain class balance during replay, generation frequencies are weighted inversely to generation counts, biased linearly by task age (see Algorithm~\ref{alg:awake}, line~\ref{line:inv_bincount}).

\subsection{Short-term memories and lateralisation}\label{sec:stm}
Each STM acts as a memory buffer storing 50 exemplars. We evaluate two buffer storage selection mechanisms:
\begin{enumerate}
    \item \textbf{Comparative Confidence Selection (CCS):} Samples from the task and prior STM are passed through the post-training LTM and ranked by confidence. In symmetric configurations, moderate-confidence quantile range samples ($\sim 50\%$) are selected in both hemispheres to represent distribution boundaries while retaining discernibility. Under asymmetric configurations, STM storage selection is also lateralised: the Left Hemisphere selects high-confidence anchor samples to reinforce representations against drift, while the Right Hemisphere selects moderate-confidence samples to explore decision boundaries. Specifically, we filter replay samples using the raw probability $\max_c \hat{y}_c = \exp(CONF(\hat{y})) \in [0,1]$, retaining candidates within the moderate-confidence quantile range $\{L, R\} = \{0.1, 0.5\}$. This design helps avoid VAE posterior collapse and representational drift.
    \item \textbf{Latent Space Cluster Centroids (LSCC):} We apply K-Means clustering in the VAE latent space across task and STM samples, retaining exemplars closest to the centroids. This grounds representations and further prevents representational drift.
\end{enumerate}

To model biological hemispheric lateralisation (stability vs. plasticity), we configure hyperparameter asymmetry. The Left Hemisphere (LH) is configured for stability and routine processing (lower generator temperature $\tau$ and task-age bias favouring older tasks). The Right Hemisphere (RH) is configured for plasticity and exploration (higher generator temperature $\tau$ and no age bias), enabling rapid adaptation to novel distributions.

\subsection{Task Transfer Metrics}\label{sec:task_transfer_metrics}

    To evaluate how sequential learning affects historical retention and future task representations, we analyse Forgetting ($F$), Backward Transfer ($\text{BWT}$), and Representation Probe
  Forward Transfer ($\text{FWT}_{\text{probe}}$).

    Forgetting ($F$) measures the average degradation in performance on past tasks relative to the maximum accuracy attained during training~\cite{chaudhry_riemannian_2018}:
    \begin{equation}
    F = \frac{1}{T-1} \sum_{i=1}^{T-1} f_i^T, \quad \text{where} \quad f_i^T = \max_{j \in \{i, \dots, T-1\}} R_{j,i} - R_{T,i}
    \end{equation}
    where $R_{j,i}$ denotes the test classification accuracy on task $i$ evaluated after completing training on task $j$, and $f_i^T \ge 0$ quantifies the drop from peak accuracy for task
  $i$ after learning all $T$ tasks.

    Backward Transfer ($\text{BWT}$) assesses how learning subsequent tasks influences historical performance relative to immediate post-task acquisition~\cite{lopez-paz_gradient_2017}:
    \begin{equation}
    \text{BWT} = \frac{1}{T-1} \sum_{i=1}^{T-1} (R_{T,i} - R_{i,i})
    \end{equation}
    where $R_{i,i}$ is the immediate test accuracy on task $i$ right after its training phase. Because the maximum historical accuracy cannot fall below the immediate post-learning accuracy
  ($\max_{j \ge i} R_{j,i} \ge R_{i,i}$), it holds analytically that $F \ge -\text{BWT}$. When past task performance reaches its peak immediately upon task acquisition and monotonically
  degrades during subsequent tasks ($\max_{j} R_{j,i} = R_{i,i}$), this relation holds with equality: $F \equiv -\text{BWT}$. Conversely, when a method exhibits positive backward transfer
  ($R_{T,i} > R_{i,i}$) without prior drops, historical forgetting is zero ($F = 0.0\%$).
  
    In class-incremental learning from scratch, classical zero-shot forward transfer \cite{lopez-paz_gradient_2017} predictably hovers near $0\%$ because the model has not yet formed output
  classification heads for unseen classes. To quantify whether upstream tasks build transferable
  latent representations that facilitate future learning, we evaluate Representation Probe Forward
  Transfer ($\text{FWT}_{\text{probe}}$) \cite{chen_forgetting_2023}:
    \begin{equation}
    \text{FWT}_{\text{probe}} = \frac{1}{T-1} \sum_{i=2}^{T} (R_{i-1,i}^{\text{probe}} - \bar{b}_i)
    \end{equation}
    where $R_{i-1,i}^{\text{probe}}$ denotes the test classification accuracy on future task $i$
  obtained by fitting a linear probe on top of the frozen representations extracted after completing
  task $i-1$, and $\bar{b}_i$ denotes the dataset-level baseline probe accuracy obtained directly from untrained representations ($97.2\%$ on Split-MNIST, $84.5\%$ on Split-Fashion-MNIST, and $18.5\%$ on Split-CIFAR-100), measuring the raw linear separability headroom of the benchmark. A positive $\Delta\text{FWT}_{\text{probe}} = R_{i-1,
  i}^{\text{probe}} - \bar{b}_i$ indicates that upstream task learning has enriched feature
  representations for downstream categories.

\subsection{Code Availability and Benchmark Reproduction}\label{sec:github}
    To ensure full experimental transparency and reproducibility, all source code, model
  configurations, and benchmark implementations are open-source and publicly hosted at
  \url{https://github.com/n3u7r1n0s/in-two-minds}. The repository provides:
    \begin{enumerate}
        \item A standalone PyTorch implementation of the 4MAS awake/sleep bilateral architecture;
        \item An integration plugin for the MAMMOTH continual learning benchmark framework;
        \item Capacity-scaled comparative baselines (including parameter-matched Brain-Inspired
  Replay);
        \item An automated reproduction script (\texttt{reproduce\_table1.py}) that reproduces all 26
  Class-IL configurations and generates the metrics reported in Table~\ref{tab:dataset_performance}.
    \end{enumerate}
\section{Results and discussion}\label{sec:results}

Our results show that 4MAS achieves competitive Class-IL accuracies across all benchmarks. As shown in Table~\ref{tab:dataset_performance}, 4MAS demonstrates strong resilience to catastrophic forgetting. On Split-MNIST, 4MAS with CCS achieves $98.30 \pm 0.04\%$ accuracy, which matches the capacity-scaled offline Joint Training ceiling of $98.14 \pm 0.12\%$ within statistical confidence bounds. This confirms that bilateral consolidation and confidence-routed ensembling virtually eliminate catastrophic forgetting on this benchmark. On Split-Fashion-MNIST, 4MAS with LSCC achieves $84.9 \pm 0.3\%$, representing a significant advancement over B-IR ($74.6 \pm 0.4\%$) and Generative Replay (GR) ($74.5 \pm 0.4\%$) \cite{shin_continual_2017}. This performance boost is directly tied to our low Forgetting scores ($1.2\%$ on Split-MNIST, $10.6\%$ on Split-Fashion-MNIST) and low representational drift ($1.0$ on both Split-MNIST and Split-Fashion-MNIST, compared to GR's $42.1\pm 1.8$ and $48.5 \pm 2.3$). This demonstrates that the dual-hemisphere and sleep-phase architecture stabilises internal latent representations, preventing the drift that typically destabilises single-generator networks.

On the challenging Split-CIFAR-100 dataset and when scaling each architecture to $\sim$140M trainable parameters, 4MAS with asymmetric CCS improves on the unilateral B-IR baseline ($27.85 \pm 0.55\%$) and significantly outperforms other generative replay methods. The performance gap relative to Joint training ($51.9 \pm 0.4\%$) is primarily due to the expressive capacity of the standard flat VAE decoder, which struggles to reconstruct high-frequency details for 100 complex classes.

Every regularisation method performs at or near the fine-tuning lower bound: parameter regularisation (EWC, oEWC and SI) never exceeds it by more than $1.7$ points on any benchmark, and functional regularisation (LwF) by at most $4.5$. Constraint-based stability alone did not support Class-IL. LwF's performance on Split-CIFAR-100 is less clear; our runs achieved $10.8 \pm 0.2\%$ and we were unable to reproduce the accuracy of $25.57\%$ reported in \cite{vandeven_three_2022} using the same reference implementation and protocol.

We note that while distinct continual learning paradigms---such as methods relying on large uncompressed exemplar buffers or pre-trained foundation models—can achieve higher absolute accuracy on CIFAR-100, 4MAS is designed to investigate task-agnostic biological memory consolidation. Within the class of generative replay and brain-inspired dual-memory architectures evaluated under matching feature protocols, 4MAS outperforms unilateral baselines and exhibits monotonic parameter scaling.

\subsection{Task Transfer Analysis}

\paragraph{Backward Transfer Dynamics}
Table~\ref{tab:dataset_performance} demonstrates that 4MAS achieves the least negative BWT among methods that successfully learn the task sequence. On Split-MNIST, Split-Fashion-MNIST, and Split-CIFAR-100, 4MAS achieves BWT scores of $-1.2\%$, $-10.6\%$, and $-25.8\%$ respectively, substantially less negative than B-IR ($-6.4\%$, $-30.2\%$ and $-52.9\%$) and GR ($-8.9\%$, $-25.4\%$ and $-66.1\%$). LwF records the least negative BWT on every benchmark ($+0.1\%$, $+0.3\%$ and $-3.9\%$), but only because it barely acquires new classes (final accuracy of at most $24.2\%$); its stability comes at the cost of plasticity, so BWT alone does not indicate successful continual learning. 4MAS therefore largely preserves past decision boundaries during new class assimilation, including on the challenging Split-CIFAR-100 benchmark, where it reduces forgetting by more than half relative to B-IR (BWT of $-25.8\%$ vs.\ $-52.9\%$).

\paragraph{Forward Transfer Dynamics ($\text{FWT}_{\text{probe}}$)}
To quantify whether earlier task learning enriches representation quality for subsequent tasks, we evaluate Representation Probe Transfer ($\text{FWT}_{\text{probe}}$) via linear probes trained on frozen representations prior to task training \cite{chen_forgetting_2023}, as seen in Table~\ref{tab:dataset_performance}. On Split-Fashion-MNIST, all architectures exhibit strong positive forward transfer ($\Delta\text{FWT}_{\text{probe}} \approx +11.1\%$ to $+12.8\%$ above random initialisation), confirming that low-level visual features (edges, contours, silhouettes) are successfully reused across garment classes. On Split-MNIST, the architectures exhibit little or no positive forward transfer ($\Delta\text{FWT}_{\text{probe}} \approx -5.7\%$ to $+1.5\%$ above random initialisation). Because MNIST digits are nearly linearly separable, linear probes on untrained random features already reach ${\sim}97\%$ accuracy, leaving little headroom for representation reuse. On Split-CIFAR-100, 4MAS and B-IR achieve substantial positive representation transfer ($\Delta\text{FWT}_{\text{probe}} = +52.3\%$ and $+51.2\%$ above untrained random initialisation). However, LwF, which uses no replay, achieves comparable transfer ($+49.3\%$), while GR achieves only $+38.2\%$. As 4MAS and B-IR are also the largest models evaluated ($\sim$140M versus 11M parameters), this advantage cannot be attributed to generative replay alone. Overall, forward transfer is driven more by the benchmark than by the continual learning method: on Split-Fashion-MNIST, all methods fall within $1.7$ points of each other. The advantage of 4MAS therefore lies in retention rather than forward transfer, and on Split-Fashion-MNIST and Split-CIFAR-100 its consolidation achieves this without reducing forward transfer.

    \begin{table*}[!t]
    \centering
    \caption{\label{tab:dataset_performance}Class-IL evaluation across Split-MNIST, Split-Fashion-MNIST, and Split-CIFAR-100 benchmarks. For each method, we report performance across capacity sweeps, with total trainable parameter counts indicated in parentheses. Best values (excluding the Joint upper bound) are in bold. Metrics display mean and standard error of the mean ($\pm\text{SEM}$) across random seeds after learning all tasks. $\text{FWT}_{\text{probe}}$ denotes representation linear probe accuracy on future tasks and relative transfer ($\Delta$) above random initialisation.}
    \resizebox{\textwidth}{!}{
    \begin{tabular}{@{}llccccc@{}}
    \toprule
    \textbf{Dataset} & \textbf{Method} & \textbf{Final Acc} & \textbf{Forg.} & \textbf{BWT} & \textbf{FWT$_{\text{probe}}$ ($\Delta$)} & \textbf{Drift} \\
    & & \textbf{(\%) $\uparrow$} & \textbf{(\%) $\downarrow$} & \textbf{(\%) $\uparrow$} & \textbf{($\%$) $\uparrow$} & \textbf{$\downarrow$} \\
    \midrule
    \multirow{9}{*}{Split-MNIST}
    & Joint (145M) & $98.14 \pm 0.12$ & — & — & — & — \\
    & \cellcolor{oursgreen} 4MAS (Our method, 145M) & \cellcolor{oursgreen} \best{$98.30 \pm 0.04$} & \cellcolor{oursgreen} $1.2$ & \cellcolor{oursgreen} $-1.2$ & \cellcolor{oursgreen} $91.5\;(-5.7)$ & \cellcolor{oursgreen} $1.0$ \\
    & B-IR (5M) & $93.5 \pm 0.2$ & $6.4 \pm 0.2$ & $-6.4 \pm 0.2$ & \best{$98.7\;(+1.5)$} & \best{$0.2 \pm 0.0$} \\
    & GR (186M) & $94.50 \pm 0.49$ & $8.9 \pm 0.4$ & $-8.9 \pm 0.4$ & $97.6\;(+0.5)$ & $42.1 \pm 1.8$ \\
    & LwF (1.9M) & $24.2 \pm 0.4$ & \best{$0.0$} & \best{$0.1$} & $96.7\;(-0.5)$ & $15.2$ \\
    & EWC (1.9M) & $19.9 \pm 0.0$ & $79.2$ & $-79.2$ & $97.1\;(-0.1)$ & $61.3$ \\
    & oEWC (1.9M) & $19.9 \pm 0.0$ & $79.2$ & $-79.2$ & $97.1\;(-0.1)$ & $58.1$ \\
    & SI (1.9M) & $19.9 \pm 0.0$ & $79.3$ & $-79.3$ & $94.3\;(-2.9)$ & $33.5$ \\
    & Fine-Tuning (1.9M) & $19.7 \pm 0.1$ & $79.5 \pm 0.1$ & $-79.5 \pm 0.1$ & $92.9\;(-4.3)$ & $84.2 \pm 2.8$ \\
    \midrule
    \multirow{9}{*}{Split-Fashion-MNIST}
    & Joint (78M) & $88.87 \pm 0.18$ & — & — & — & — \\
    & \cellcolor{oursgreen} 4MAS (Our method, 145M) & \cellcolor{oursgreen} \best{$84.9 \pm 0.3$} & \cellcolor{oursgreen} $10.6$ & \cellcolor{oursgreen} $-10.6$ & \cellcolor{oursgreen} $96.5\;(+12.0)$ & \cellcolor{oursgreen} $1.0$ \\
    & B-IR (11M) & $74.6 \pm 0.4$ & $30.2 \pm 0.3$ & $-30.2 \pm 0.3$ & \best{$97.3\;(+12.8)$} & \best{$0.2 \pm 0.0$} \\
    & GR (11M) & $74.53 \pm 0.43$ & $25.4 \pm 0.3$ & $-25.4 \pm 0.3$ & $96.8\;(+12.3)$ & $48.5 \pm 2.3$ \\
    & LwF (1.2M) & $20.1 \pm 0.2$ & \best{$0.0$} & \best{$0.3$} & $95.8\;(+11.3)$ & $12.9$ \\
    & EWC (1.2M) & $19.9 \pm 0.0$ & $79.2$ & $-79.2$ & $96.4\;(+11.9)$ & $55.5$ \\
    & oEWC (1.2M) & $19.8 \pm 0.4$ & $79.2$ & $-79.2$ & $96.3\;(+11.8)$ & $50.3$ \\
    & SI (1.2M) & $19.9 \pm 0.0$ & $79.3$ & $-79.3$ & $96.1\;(+11.6)$ & $30.4$ \\
    & Fine-Tuning (1.2M) & $20.0 \pm 0.0$ & $79.4 \pm 0.0$ & $-79.4 \pm 0.0$ & $95.6\;(+11.1)$ & $77.3 \pm 2.3$ \\
    \midrule
    \multirow{8}{*}{Split-CIFAR-100}
    & Joint (35M) & $51.9 \pm 0.4$ & — & — & — & — \\
    & \cellcolor{oursgreen} 4MAS (Our method, 140M) & \cellcolor{oursgreen} \best{$29.29 \pm 0.29$} & \cellcolor{oursgreen} $25.8 \pm 1.2$ & \cellcolor{oursgreen} $-25.8 \pm 1.2$ & \cellcolor{oursgreen} \best{$70.8\;(+52.3)$} & \cellcolor{oursgreen} \best{$0.43 \pm 0.07$} \\
    & B-IR (140M) & $27.85 \pm 0.55$ & $52.9$ & $-52.9$ & $69.7\;(+51.2)$ & $5.1$ \\
    & LwF (11M) & $10.8 \pm 0.2$ & \best{$4.6 \pm 0.3$} & \best{$-3.9 \pm 0.3$} & $67.8\;(+49.3)$ & $1.9$ \\
    & SI (11M) & $9.39 \pm 0.27$ & $79.4 \pm 0.4$ & $-79.4 \pm 0.4$ & $55.4\;(+36.9)$ & $3.1$ \\
    & GR (11M) & $7.9 \pm 0.1$ & $66.1 \pm 0.5$ & $-66.1 \pm 0.5$ & $56.7\;(+38.2)$ & $10.3$ \\
    & Fine-Tuning (11M) & $7.73 \pm 0.15$ & $81.5 \pm 0.3$ & $-81.5 \pm 0.3$ & $53.5\;(+35.0)$ & $3.6$ \\
    & EWC (11M) & $8.1 \pm 0.1$ & $81.6 \pm 0.4$ & $-81.6 \pm 0.4$ & $52.7\;(+34.2)$ & $4.2$ \\
    \bottomrule
    \end{tabular}
    }
    \vspace{4pt}
    \begin{minipage}{\textwidth}
      \scriptsize \textbf{Model Definitions \& Citations:}
      \textbf{4MAS}: 4 Module Awake/Sleep (Our method); \textbf{B-IR}: Brain-Inspired Replay~\cite{vandeven_braininspired_2020}; \textbf{GR}: Generative Replay~\cite{shin_continual_2017}; \textbf{LwF}: Learning without Forgetting~\cite{li_learning_2018}; \textbf{EWC}: Elastic Weight Consolidation~\cite{kirkpatrick_overcoming_2017}; \textbf{oEWC}: online Elastic Weight Consolidation~\cite{schwarz_progress_2018}; \textbf{SI}: Synaptic Intelligence~\cite{zenke_continual_2017}; \textbf{Joint}: Offline training upper bound; \textbf{Fine-Tuning}: Sequential learning without memory or regularisation. Dashes (—) for Joint Training across all benchmarks indicate Forgetting, BWT, FWT, and Drift are not applicable because all classes are trained concurrently in a single non-sequential offline ceiling stage. Forgetting is measured as the drop from maximum historical accuracy~\cite{chaudhry_riemannian_2018}. For LwF on Split-MNIST and Split-Fashion-MNIST, positive backward transfer ($+0.1\%, +0.3\%$) results in $0.0\%$ forgetting as accuracy does not drop below initial post-task levels.
    \end{minipage}
    \end{table*}
\subsection{Sleep tuning}
\begin{figure}
\centering
    \begin{minipage}[t]{0.47\textwidth}
        \centering
        \textbf{(a)} \\
        \includegraphics[width=\linewidth]{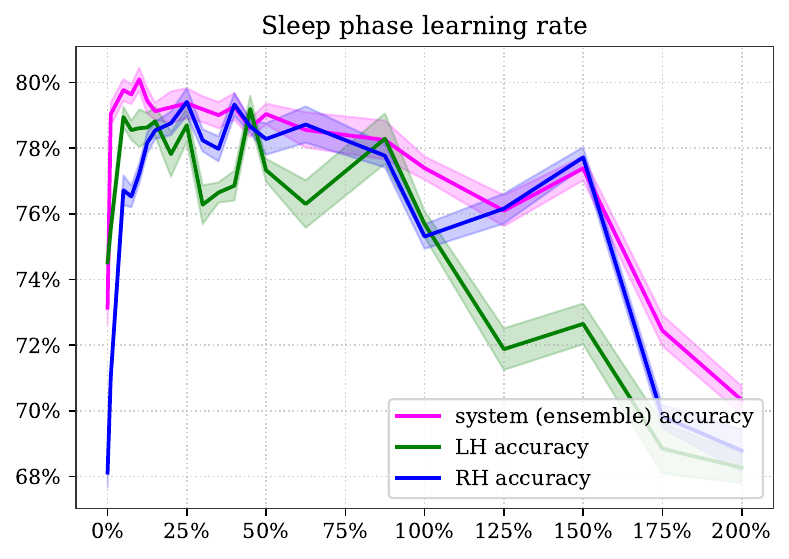}
    \end{minipage}
    \hfill
    \begin{minipage}[t]{0.47\textwidth}
        \centering
        \textbf{(b)} \\
        \includegraphics[width=\linewidth]{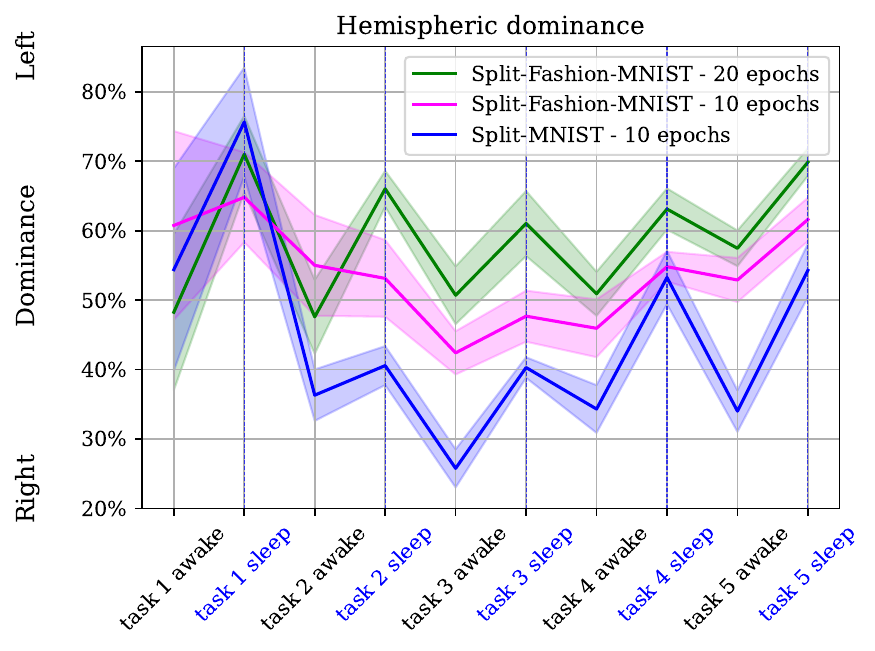}
    \end{minipage}
  \caption{\label{fig:sleep_lr}(a) Performance variation with sleep phase learning rates (LR) as a percentage of awake phase LR, Split-Fashion-MNIST. (b) Dominance scores categorised as RH (<50\%), symmetrical (50\%), or LH (>50\%). Memory consolidation during sleep phase increased LH dominance, while new task learning during awake phase increased RH dominance; produced by lateralisation techniques (class representation bias by age in LH and asymmetric generator temperatures).}
\end{figure}

We evaluated sleep phase learning rates ($LR$) ranging from 0\% to 200\% of the awake phase rate (Figure~\ref{fig:sleep_lr}a). On Split-Fashion-MNIST a reduced sleep rate of 5–10\% ($\approx 2 \times 10^{-5}$) performed best ($84.9 \pm 0.3\%$), while removing sleep ($LR=0\%$) yielded $74.0 \pm 0.59\%$. Larger learning rates degraded performance, confirming that low sleep learning rates enable fine-tuning on the contralateral representation without overwriting specialised features.

Furthermore, average ensemble accuracy exceeded either isolated hemisphere. This gain stems from effective confidence-based routing between specialised hemispheres rather than standard ensemble variance reduction; indeed, without sleep-phase consolidation, the ensemble fails to outperform the strongest individual hemisphere (Section~\ref{sec:ablation}). Because the router selects whichever hemisphere is more confident at inference, the share of inferences each one wins measures its dominance, and this shifted dynamically over training: wake training on new tasks increased RH dominance, while sleep consolidation restored LH dominance (Figure~\ref{fig:sleep_lr}b). This shift aligns with Goldberg's Novelty-Routine hypothesis, reflecting a transition from initial RH-driven processing of novel representations to consolidated, routinised LH schemas \cite{goldberg_new_2018}.

\subsection{Lateralisation}
\begin{figure}
\centering
    \begin{minipage}[t]{0.47\textwidth}
        \centering
        \textbf{(a)} \\
        \includegraphics[width=\linewidth]{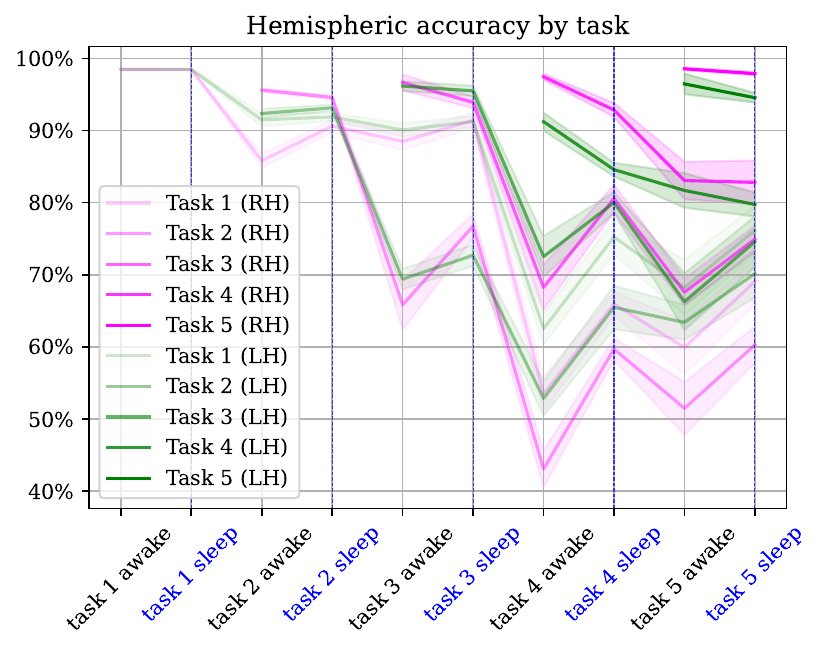}
    \end{minipage}
    \hfill
    \begin{minipage}[t]{0.47\textwidth}
        \centering
        \textbf{(b)} \\
        \includegraphics[width=\linewidth]{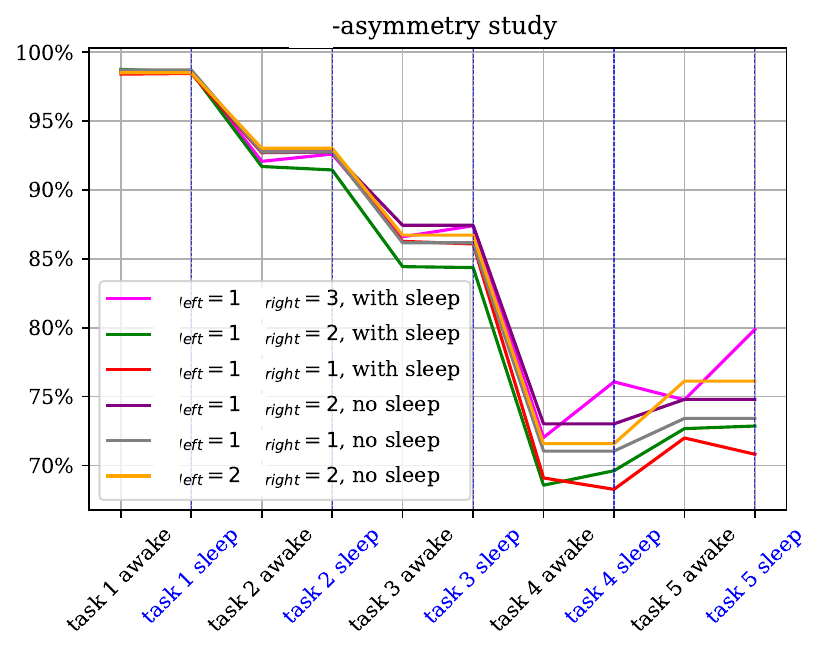}
    \end{minipage}
  \caption{\label{fig:asymmetry}Fashion-MNIST: (a) Hemispheric accuracy for each task. (b) $\tau$-asymmetry results, with and without sleep training phase.}
\end{figure}

Figure~\ref{fig:asymmetry}a shows that the LH accuracy decays slowly, maintaining stability, while the RH learns new tasks quickly but forgets faster, supporting the stability-plasticity lateralisation described in the Novelty-Routine hypothesis \cite{goldberg_new_2018}.
Lateralisation of the generator temperature ($\tau$) significantly improved performance (Table~\ref{tab:gen_temps}, Figure~\ref{fig:asymmetry}b). Symmetrical temperatures ($\tau=1$ or $\tau=2$ for both) caused degraded performance or limited sleep benefits, whereas asymmetrical configurations ($\tau=1$ for LH, $\tau \in [2,4]$ for RH) showed consistent performance gains after each sleep phase. This indicates that specialisation, elicited by asymmetrical parameterisation, provides an advantage and better use of total resources.

\begin{figure}
\centering
    \begin{minipage}[t]{0.47\textwidth}
        \centering
        \textbf{(a)} \\
        \includegraphics[width=\linewidth]{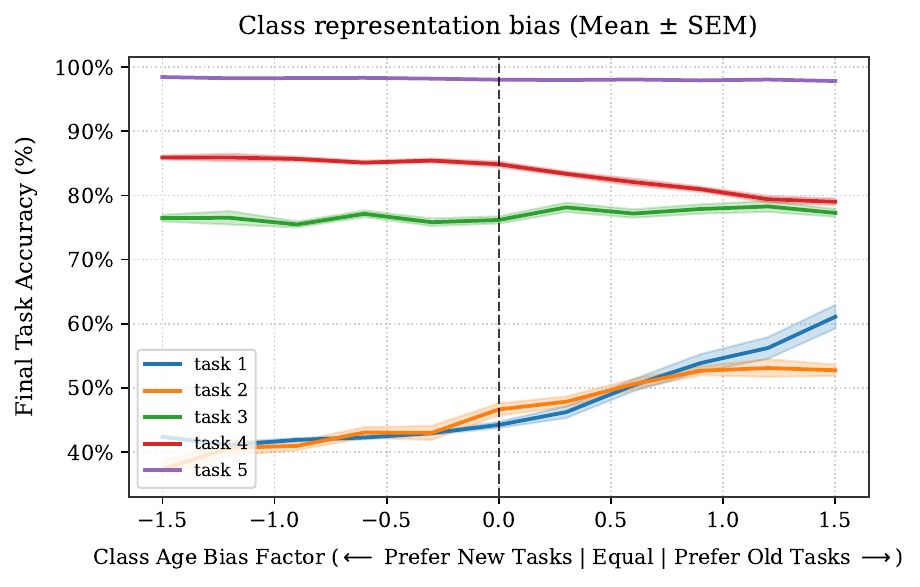}
    \end{minipage}
    \hfill
    \begin{minipage}[t]{0.47\textwidth}
        \centering
        \textbf{(b)} \\
        \includegraphics[width=\linewidth]{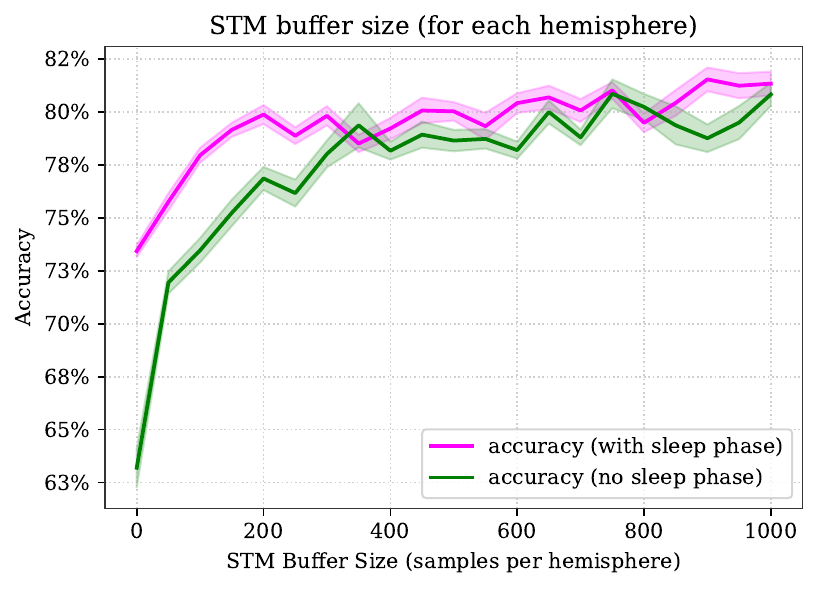}
    \end{minipage} 
\caption{\label{fig:class_bias}(a) Biasing class representation by recency; weighting training heavily toward older classes significantly boosts early-class accuracy with only minor performance degradation on recent classes. (b) Effect of STM memory buffer size on accuracy; gains from larger buffers plateau after 200 samples, while also diminishing the relative impact of the sleep phase.}
\end{figure}

Applying a task-age class representation bias to the LH only (biasing training toward older tasks) improved the retention of earlier tasks by 15–26\% (Figure~\ref{fig:class_bias}a), allowing the ensemble to retain stability in LH while maintaining plasticity in RH.

\begin{table}
\centering
\begin{tabular}{|c|c|c|c|c|c|c|}
\hline
\multicolumn{2}{|c|}{} & \multicolumn{5}{c|}{$\tau$ - right} \\
\cline{3-7}
\multicolumn{2}{|c|}{} & \textbf{1} & \textbf{2} & \textbf{3} & \textbf{4}  & \textbf{5} \\
\hline
\multirow{5}{*}{\rotatebox{90}{$\tau$ - left}} & \textbf{1} & \cellcolor{heatmap4}0.6301 & \cellcolor{heatmap8}0.7122 & \cellcolor{heatmap9}\best{0.7445} & \cellcolor{heatmap8}0.7132 & \cellcolor{heatmap7}0.6984 \\
\cline{2-7}
& \textbf{2} & \cellcolor{heatmap7}0.7021 & \cellcolor{heatmap6}0.6895 & \cellcolor{heatmap5}0.6662 & \cellcolor{heatmap4}0.6342 & \cellcolor{heatmap3}0.6178 \\
\cline{2-7}
& \textbf{3} & \cellcolor{heatmap6}0.6768 & \cellcolor{heatmap3}0.6116 & \cellcolor{heatmap2}{0.5907} & \cellcolor{heatmap1}0.5595 & \cellcolor{heatmap1}0.5504 \\
\cline{2-7}
& \textbf{4} & \cellcolor{heatmap6}0.6593 & \cellcolor{heatmap3}0.6128 & \cellcolor{heatmap1}0.5561 & \cellcolor{heatmap1}0.5551 & \cellcolor{heatmap0}0.5233 \\
\cline{2-7}
& \textbf{5} & \cellcolor{heatmap5}0.6453 & \cellcolor{heatmap2}0.576 & \cellcolor{heatmap0}0.5417 & \cellcolor{heatmap1}0.555 & \cellcolor{heatmap0}0.5148 \\
\hline
\end{tabular}
  \caption{\label{tab:gen_temps}LH and RH generator temperature vs Split-Fashion-MNIST accuracy. Degraded performance observed for symmetrical temperatures.}
\end{table}

\subsection{STM buffer selection}
Evaluating CCS and LSCC selection mechanisms (Table~\ref{tab:centroids}) showed that storing moderate-confidence samples ($\sim 50\%$) under CCS performed best, whereas selecting low-confidence outliers led to VAE posterior collapse. Storing latent space cluster centroids via K-Means (LSCC) yielded the best and most consistent results across MNIST and Fashion-MNIST tasks, indicating that the STM buffer is most effective when its primary role is grounding the latent space against representational drift rather than importing weakly-learned outliers. Since the STM is the source of the exemplars replayed in both the awake and sleep phases, what is stored determines what is later reactivated. This parallels neurobiological evidence that offline hippocampal reactivation is selective: which memories are recruited predicts the stabilisation of cognitive maps that comprehensively reflect the entire spatial context \cite{grosmark_reactivation_2021}. In both cases the benefit depends on what is chosen for reactivation, not on reactivation alone.

The robust performance of the asymmetric threshold configuration ($L=0.1, R=0.5$) on CIFAR-100 is explained by a functional division of labour. The left hemisphere, operating at a low generator temperature ($\tau=1$), behaves as a stable anchor that preserves core, high-confidence representations. By setting a very conservative threshold ($L=0.1$), we prevent representational drift during consolidation. Conversely, the right hemisphere, operating at a high generator temperature ($\tau=3$), acts as a flexible explorer. Storing intermediate-confidence boundary samples ($R=0.5$) allows it to explore variations and refine task boundaries, leading to significantly enhanced ensemble consolidation.

\begin{table}
\centering
\scriptsize
\begin{tabular}{|c|c|c|c|c|c|c|c|c|}
\hline
\multicolumn{2}{|c|}{} & \multicolumn{6}{c|}{CCS - RH} & \\
\cline{3-9}
\multicolumn{2}{|c|}{} & \textbf{90\%} & \textbf{70\%} & \textbf{50\%} & \textbf{30\%}  & \textbf{10\%}  & \textbf{0\%} & \textbf{LSCC}\\
\hline
\multirow{6}{*}{\rotatebox{90}{CCS - LH}} & \textbf{90\%} & \cellcolor{heatmap5}0.7562 & \cellcolor{heatmap6}0.7662 & \cellcolor{heatmap6}0.7668 & \cellcolor{heatmap7}0.7668 & \cellcolor{heatmap5}0.7536 & \cellcolor{heatmap4}0.7505 & \cellcolor{heatmap6}0.7656 \\
\cline{2-9}
& \textbf{70\%} & \cellcolor{heatmap6}0.7618 & \cellcolor{heatmap7}0.7724 & \cellcolor{heatmap8}0.7753 & \cellcolor{heatmap7}0.7725 & \cellcolor{heatmap6}0.7607 & \cellcolor{heatmap5}0.7589 & \cellcolor{heatmap8}0.7796 \\
\cline{2-9}
& \textbf{50\%} & \cellcolor{heatmap7}0.7695 & \cellcolor{heatmap7}0.7706 & \cellcolor{heatmap8}{0.7733} & \cellcolor{heatmap7}0.7667 & \cellcolor{heatmap6}0.7648 & \cellcolor{heatmap5}0.7573 & \cellcolor{heatmap8}0.7779 \\
\cline{2-9}
& \textbf{30\%} & \cellcolor{heatmap7}0.7703 & \cellcolor{heatmap7}0.7718 & \cellcolor{heatmap8}0.7758 & \cellcolor{heatmap6}0.7627 & \cellcolor{heatmap5}0.7565 & \cellcolor{heatmap0}0.7247 & \cellcolor{heatmap9}\best{0.7799} \\
\cline{2-9}
& \textbf{10\%} & \cellcolor{heatmap7}0.7673 & \cellcolor{heatmap7}0.7704 & \cellcolor{heatmap7}0.7675 & \cellcolor{heatmap4}0.7509 & \cellcolor{heatmap2}0.7351 & \cellcolor{heatmap0}0.727 & \cellcolor{heatmap8}0.7757 \\
\cline{2-9}
& \textbf{0\%} & \cellcolor{heatmap5}0.7547 & \cellcolor{heatmap7}0.7726 & \cellcolor{heatmap6}0.7613 & \cellcolor{heatmap4}0.7470 & \cellcolor{heatmap1}0.7285 & \cellcolor{heatmap0}0.7205 & \cellcolor{heatmap6}0.7662 \\
\cline{1-9}
& \textbf{LSCC} & \cellcolor{heatmap8}0.7748 & \cellcolor{heatmap8}0.7776 & \cellcolor{heatmap9}\best{0.7799} & \cellcolor{heatmap8}0.7764 & \cellcolor{heatmap6}0.7607 & \cellcolor{heatmap5}0.7588 & \cellcolor{heatmap8}0.776 \\
\hline
\end{tabular}
  \caption{\label{tab:centroids}Comparison of mechanisms for selecting Fashion-MNIST memories for STM buffer storage. Storing latent space centroids via K-Means clustering (LSCC) produced the most consistent results on Split-Fashion-MNIST, whereas highly asymmetric confidence selection ($L=0.1/R=0.5$) (CCS) achieved peak performance on the more complex Split-CIFAR-100 benchmark.}
\end{table}

\paragraph{Asymmetry as recurring mechanism.}
The performance of 4MAS depends less on any individual component than on the interaction between long and short term memories and between two hemispheres that are configured differently. Sleep learning is a fraction of the awake learning rate, so that consolidation refines rather than overwrites; the generators run at different temperatures, the LH anchoring stable representations while the RH explores; and the STM thresholds differ, the LH storing only high-confidence samples while the RH stores intermediate-confidence boundary samples. Generator temperature shows this most directly: the peak asymmetric setting ($\tau=1$ for LH, $\tau=3$ for RH) exceeds the best symmetric setting by $5.5$ points on Split-Fashion-MNIST. Asymmetric confidence thresholds are likewise best on Split-CIFAR-100, though symmetric LSCC selection wins on Split-Fashion-MNIST (Table~\ref{tab:centroids}). Specialisation between hemispheres therefore helps, but asymmetry alone is not sufficient.

\subsection{Capacity and architectural scaling}\label{sec:scaling}
\begin{figure*}[t]
\centering
\includegraphics[width=\linewidth]{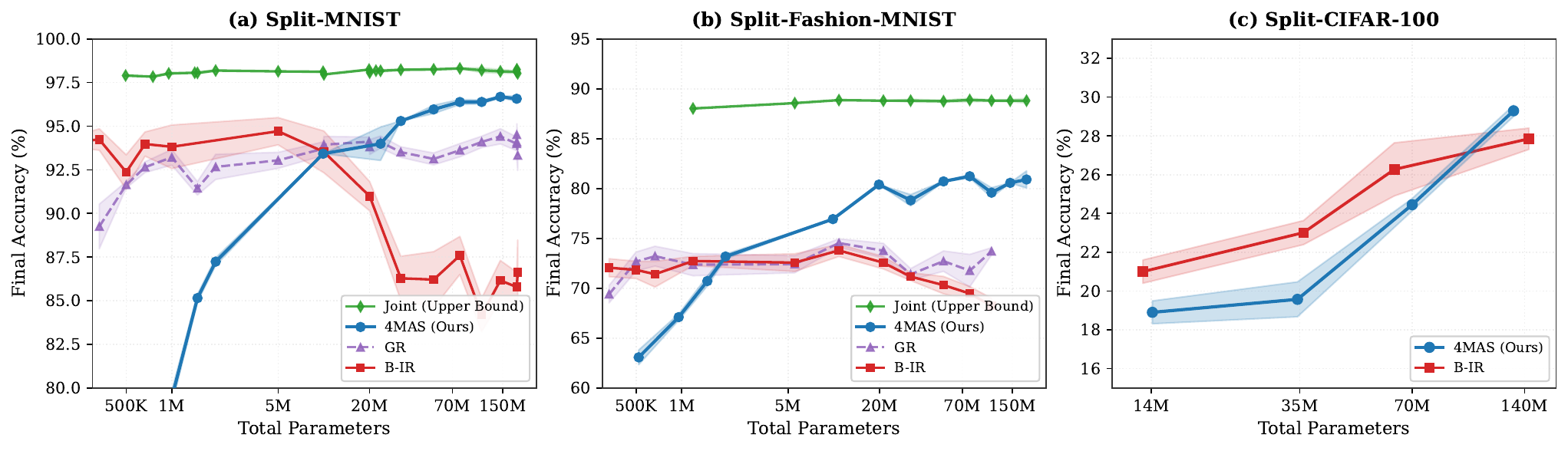}
  \caption{\label{fig:capacity_scaling}Capacity scaling curves comparing our bilateral 4MAS architecture against the unilateral B-IR baseline across (a) Split-MNIST, (b) Split-Fashion-MNIST, and (c) Split-CIFAR-100. The plots show final accuracy as a function of total model parameters (in millions).}
\end{figure*}

To evaluate how model capacity influences continual learning performance, we systematically compared the parameter scaling behaviour of 4MAS against the unilateral B-IR baseline across Split-MNIST, Split-Fashion-MNIST, and Split-CIFAR-100 (Figure~\ref{fig:capacity_scaling}). 

At lower parameter ranges (e.g., $<30$\,M parameters), monolithic generative models demonstrate superior sample and parameter efficiency. This is primarily because 4MAS splits its total parameter budget across two distinct hemispheric models (LH and RH) and requires sleep-phase cross-replay to consolidate knowledge, introducing an architectural capacity overhead. When the overall parameter budget is highly constrained, the split-hemisphere bottleneck limits the representation capacity of the individual generators. 

However, 4MAS demonstrates superior scalability as model capacity increases. Unilateral networks typically suffer from severe representational drift and catastrophic interference when forced to represent a large number of conflicting class distributions in a single unified latent space~\cite{vandeven_three_2022}. Consequently, performance of B-IR (and other unilateral models) plateaus or degrades at larger scales. In contrast, 4MAS's bilateral hemispheric partitioning and stability-plasticity division of labour mitigate representational drift, enabling monotonic scaling. At larger parameter scales, 4MAS consistently outperforms benchmarks on the Fashion-MNIST benchmark and modestly surpasses generative replay CIFAR-100 baselines.

\subsection{Ablation studies}\label{sec:ablation}
To evaluate the contribution of each component, we performed ablations on the STM buffer size, the sleep phase, and the dual-hemisphere structure.
Varying the unilateral STM buffer size from 0 to 1,000 samples (Figure~\ref{fig:class_bias}b) showed diminishing gains beyond 200 samples, improving Split-Fashion-MNIST accuracy from 73.4\% (fully ablated STM) to 81.3\%. In the absence of a sleep phase, small buffer sizes severely degraded performance ($63.2\%$ at size 0), while larger buffers offset this loss, narrowing the sleep phase benefit to $<1\%$.

\begin{figure}
\centering
\begin{minipage}[b]{0.48\textwidth}
        \centering
        \includegraphics[width=\textwidth]{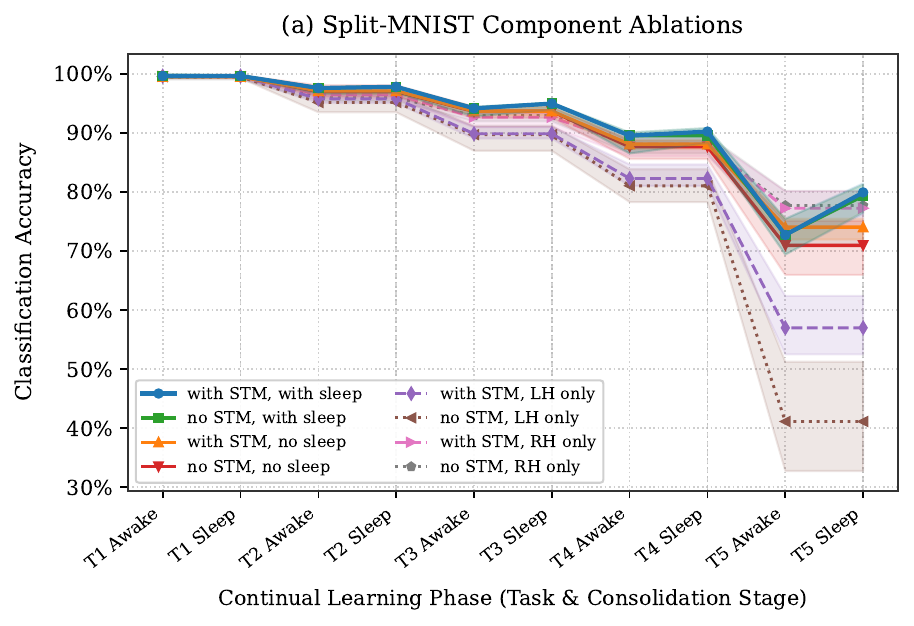}
    \end{minipage}
    \hfill
    \begin{minipage}[b]{0.48\textwidth}
        \centering
        \includegraphics[width=\textwidth]{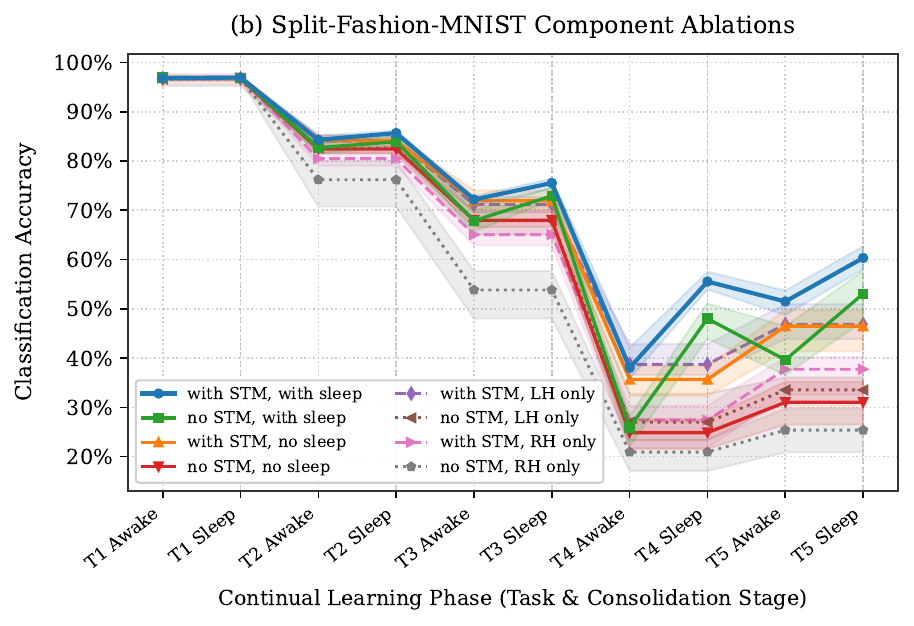}
    \end{minipage}   
\caption{\label{fig:ablations}Ablation performance for Split-MNIST and Split-Fashion-MNIST. Configurations including the sleep phase consistently perform best on both datasets, where sleep informed hemispheric confidence levels lead to system accuracy exceeding individual hemisphere accuracies.}
\end{figure}

Ablation comparisons (Figure~\ref{fig:ablations}) show that configurations including the sleep phase consistently perform best on both datasets. Without sleep, the ensemble accuracy matches its single best-performing hemisphere, confirming that sleep-based cross-replay is crucial for bilateral knowledge integration. In particular, the comparative confidence levels of each hemisphere are aligned during sleep training, allowing for system accuracy to exceed either hemisphere's individual accuracy.

\section{Limitations and future work}\label{sec:limitations}

Generative models in this study were standard non-convolutional VAEs, chosen to evaluate task-agnostic macroarchitectures. Future research should scale generators to convolutional VAEs, GANs, or diffusion models to handle more complex image distributions \cite{podoliak_development_2024,goodfellow_generative_2014,ho_denoising_2020}. Additionally, regularisation mechanisms (e.g., EWC or SI) could be integrated in tandem with our replay structure to further boost stability, as demonstrated in hybrid models like B-IR \cite{vandeven_braininspired_2020}.

While we evaluated 4MAS up to a 10-task Split-CIFAR-100 benchmark, autonomous agents in the real world require handling longer task horizons \cite{parisi_continual_2019}. Subsequent work should test 4MAS under unbounded dynamic streams \cite{khetarpal_continual_2022}, explore diverse forms of hemispheric asymmetry, and investigate additional biological mechanisms, such as slow-wave sleep modelling, to enhance persistent lifelong learning \cite{diekelmann_slowwave_2010}.

\section{Conclusion}

Lifelong learning in artificial networks remains constrained by catastrophic forgetting. Addressing this challenge, this paper introduced 4MAS, a novel continual learning macroarchitecture derived from biological memory consolidation. By modelling experience replay, bilateral sleep consolidation, and inter-hemispheric lateralisation, 4MAS employs two asymmetric LTM/STM hemispheres that coordinate via a consolidation sleep phase. Our empirical results on Split-MNIST, Split-Fashion-MNIST, and Split-CIFAR-100 demonstrate competitive performance and robust knowledge retention against representational drift. Our transfer analysis further indicates that these gains arise from retention rather than forward transfer: 4MAS reduces forgetting by more than half relative to the strongest replay baseline on Split-CIFAR-100, while its forward transfer on the more complex benchmarks remains comparable to or better than that of the baselines. Our analysis reveals that functional specialisation is essential to these gains: while symmetrical configurations degraded performance or restricted sleep benefits, our asymmetrical parameterisation consistently produced post-sleep performance enhancements, demonstrating that lateralised specialisation improves overall resource utilisation.

This work illustrates how system-level neurobiological structures can be abstracted to manage the stability-plasticity trade-off. Whereas the regularisation methods we evaluated attained stability only by sacrificing plasticity, 4MAS retained past knowledge while continuing to learn new classes. Moving beyond local weight-level constraints, system-level bilateral consolidation offers a promising path toward persistent and adaptable artificial intelligence.

\bibliographystyle{elsarticle-num} 
\bibliography{references}

\appendix
\section{Hyperparameter Searches and Optimisation}\label{sec:appendix_hyperparameters}

This appendix provides detail on the hyperparameter tuning sweeps conducted to optimise the 4MAS architecture across the three benchmarks: Split-MNIST, Split-Fashion-MNIST, and Split-CIFAR-100. These searches cover the grid-search sweeps logged in the main project database (comprising 4,328 unique trials) as well as the sequential Bayesian optimisation sweeps conducted using the Optuna framework. Hyperparameter sweeps reported in this Appendix are single runs. Code, experimental configurations, and standalone implementations are open-source and publicly accessible at \url{https://github.com/n3u7r1n0s/in-two-minds}.

\subsection{Split-MNIST and Split-Fashion-MNIST Grid Searches}\label{subsec:app_mnist_fmnist}

For both Split-MNIST and Split-Fashion-MNIST, we ran systematic parameter grids to evaluate the interaction of model size, latent representation dimensions, and generator capacity under our bilateral sleep-consolidation model. 

The range of hyperparameters explored in these grid searches is summarised in Table~\ref{tab:app_mnist_fmnist_ranges}. 

\begin{table}[h]
\centering
\caption{Hyperparameter search space for Split-MNIST and Split-Fashion-MNIST.}
\label{tab:app_mnist_fmnist_ranges}
\footnotesize
\begin{tabular}{p{4.2cm}p{6.8cm}}
\toprule
\textbf{Hyperparameter} & \textbf{Explored Values} \\
\midrule
Model Capacity (Parameters) & 248K, 477K, 736K, 970K, 4.9M, 9.9M, 11.8M, 16.0M, 26.3M, 39.2M, 54.5M, 72.4M, 92.7M \\
Short-Term Memory Size ($N_{stm}$) & 50 per hemisphere (10 samples per class) \\
VAE Generator Dimension & 140, 250, 360, 450, 1450, 2250, 2500, 3000, 4000, 5000, 6000, 7000, 8000 \\
VAE Latent Space Dimension & 16, 32, 64, 128, 256 \\
Awake-Phase Learning Rate ($LR_{awake}$) & $2 \times 10^{-4}$ (Adam) \\
Sleep-Phase Learning Rate ($LR_{sleep}$) & $2 \times 10^{-5}$ (Adam) \\
Awake-Phase Gating End-Weight & 10.0 \\
Left Generator Temperature ($\tau_L$) & 1.0 \\
Right Generator Temperature ($\tau_R$) & 3.0 \\
Oversampling Factor ($O$) & 10, 50 \\
\bottomrule
\end{tabular}
\end{table}

Large model capacity paired with a larger VAE dimension and low latent dimensionality (16) consistently yielded optimal Class-IL accuracy by preserving high-fidelity reconstructions without latent-space drift.

\subsection{Split-CIFAR-100 Optuna Optimisation Sweeps}\label{subsec:app_cifar}

Due to the increased complexity of CIFAR-100, we executed 14 distinct optimisation sweeps using the Optuna framework, focusing on architectural changes, pre-training steps, and latent alignment strategies. The best-performing trial configuration for each sweep is documented in Table~\ref{tab:app_optuna_sweeps}.

\begin{table}[h]
\centering
\caption{Best configurations and parameters across Split-CIFAR-100 Optuna sweeps.}
\label{tab:app_optuna_sweeps}
\footnotesize
\begin{tabular}{p{3.7cm}ccp{5.0cm}}
\toprule
\shortstack{\textbf{Optuna}\\\textbf{Study Name}} & \textbf{Trials} & \shortstack{\textbf{Best}\\\textbf{Class-IL (\%)}} & \shortstack{\textbf{Key Parameters}\\\textbf{and Configurations}} \\
\midrule
fourmas\_\allowbreak cifar100\_\allowbreak v1 & 89 & 0.00 & Default VAE, latent dim 64, LR 0.0014, SGD \\
fourmas\_\allowbreak cifar100\_\allowbreak v2 & 34 & - & Task-IL scenario ($28.82\%$ Task-IL), latent dim 128, buffer 2000, Adam \\
fourmas\_\allowbreak cifar100\_\allowbreak v3 & 10 & 4.11 & Class-IL, buffer 5000, VAE temp 1.15 (LH) / 1.54 (RH) \\
fourmas\_\allowbreak cifar100\_\allowbreak v4 & 26 & 7.63 & Class-IL, buffer 5000, SWS/REM epochs 75/50, LR 0.0022 \\
fourmas\_\allowbreak cifar100\_\allowbreak multi\_arch & 4 & 3.77 & Pre-trained tap layer 2 feature extractor, latent dim 64 \\
fourmas\_\allowbreak cifar100\_\allowbreak scratch\_sota & 15 & 6.85 & Scratch training (no ImageNet), buffer 10000, latent dim 128 \\
fourmas\_\allowbreak cifar100\_\allowbreak novel\_designs & 24 & 9.54 & Evaluated classifier freezing policies, distill weight 0.16 \\
fourmas\_\allowbreak cifar100\_\allowbreak refined\_sota & 17 & 9.49 & VAE dim 400, latent dim 64, gating end-weight 2.0 \\
fourmas\_\allowbreak cifar100\_\allowbreak final\_sota & 25 & 9.56 & Optimised lr 0.0023, REM/SWS epochs 150/100 \\
fourmas\_\allowbreak cifar100\_\allowbreak fixed\_sota & 13 & 9.99 & Cross-entropy loss function, linear classifier, LR 0.0018 \\
fourmas\_\allowbreak cifar100\_\allowbreak sota\_push & 34 & 11.06 & Latent dim 192, gating end-weight 1.5, LR 0.0019 \\
fourmas\_\allowbreak cifar100\_\allowbreak sota\_breakthrough\_v2 & 25 & 9.56 & VAE dim 800, latent dim 192, gating end-weight 1.0 \\
fourmas\_\allowbreak cifar100\_\allowbreak pretrain\_multi\_task & 16 & \textbf{12.25} & Bootloaded multi-task pre-training (1 task), latent dim 192 \\
fourmas\_\allowbreak cifar100\_\allowbreak vqvae & 62 & 10.35 & Vector Quantized VAE, gating end-weight 2.0, LR 0.0017 \\
fourmas\_\allowbreak cifar100\_\allowbreak vqvae\_high\_cap & 32 & 9.47 & High capacity VQ-VAE, codebook size 2048, latent dim 128 \\
\bottomrule
\end{tabular}
\end{table}

\subsection{Split-CIFAR-100 Architectural Grid Search}\label{subsec:app_cifar_grid}

To clarify the contribution of supervised contrastive loss, left-hemisphere parameter freezing, and bilateral sleep-consolidation (BSS) active gating, we conducted a 16-configuration grid search on Split-CIFAR-100. The complete experimental results of this grid search are summarised in Table~\ref{tab:app_cifar_grid}.

\begin{table}[h]
\centering
\caption{Grid search performance over generative model types, supervised contrastive (SupCon) weights, left-hemisphere parameter freezing, and BSS active gating.}
\label{tab:app_cifar_grid}
\footnotesize
\begin{tabular}{ccccccc}
\toprule
\textbf{ID} & \shortstack{\textbf{Model}\\\textbf{Type}} & \shortstack{\textbf{SupCon}\\\textbf{Weight}} & \shortstack{\textbf{LH}\\\textbf{Freeze}} & \shortstack{\textbf{BSS}\\\textbf{Active}} & \shortstack{\textbf{Class-IL}\\\textbf{(\%)}} & \shortstack{\textbf{Task-IL}\\\textbf{(\%)}} \\
\midrule
1 & vae & 0.0 & False & False & 1.52 & 12.04 \\
2 & vae & 0.0 & False & True & 1.02 & 9.61 \\
3 & vae & 0.0 & True & False & 1.19 & 11.17 \\
4 & vae & 0.0 & True & True & 1.13 & 9.26 \\
5 & vae & 0.5 & False & False & \textbf{2.13} & \textbf{17.33} \\
6 & vae & 0.5 & False & True & 2.01 & 14.90 \\
7 & vae & 0.5 & True & False & 1.03 & 10.32 \\
8 & vae & 0.5 & True & True & 1.06 & 10.51 \\
9 & vqvae & 0.0 & False & False & 1.41 & 10.31 \\
10 & vqvae & 0.0 & False & True & 1.07 & 14.19 \\
11 & vqvae & 0.0 & True & False & 1.11 & 10.45 \\
12 & vqvae & 0.0 & True & True & 1.08 & 9.38 \\
13 & vqvae & 0.5 & False & False & 1.52 & 12.41 \\
14 & vqvae & 0.5 & False & True & 1.00 & 10.87 \\
15 & vqvae & 0.5 & True & False & 1.04 & 10.47 \\
16 & vqvae & 0.5 & True & True & 1.01 & 9.96 \\
\bottomrule
\end{tabular}
\end{table}

\subsection{Split-CIFAR-100 Ablation and Asymmetry Trials}\label{subsec:app_cifar_ablation}

To systematically investigate the roles of comparative confidence selection (CCS), latent space cluster centroids (LSCC), pre-training tasks (1 vs. 5), and synaptic intelligence (SI) regularisation on the Split-CIFAR-100 benchmark, we evaluated several design configurations. The results are summarised in Table~\ref{tab:app_cifar_ablation_results}.

\begin{table}[h]
\centering
\caption{Split-CIFAR-100 performance across selection strategies (CCS, LSCC, Hybrid), pre-training task horizons, and local SI regularisation.}
\label{tab:app_cifar_ablation_results}
\resizebox{\linewidth}{!}{%
\begin{tabular}{lccccc}
\toprule
\textbf{Selection Strategy (LH / RH)} & \textbf{Left Centroid ($L$)} & \textbf{Right Centroid ($R$)} & \shortstack{\textbf{Pre-training}\\\textbf{Tasks}} & \shortstack{\textbf{Synaptic}\\\textbf{Intelligence (SI)}} & \shortstack{\textbf{Class-IL}\\\textbf{Acc (\%)}} \\
\midrule
CCS (Symmetric Baseline) & 0.5 & 0.5 & 1 & No & 21.38 \\
CCS (Symmetric) & 0.5 & 0.5 & 2 & No & 25.34 \\
CCS (Symmetric) & 0.5 & 0.5 & 3 & No & 29.18 \\
CCS (Symmetric) & 0.5 & 0.5 & 4 & No & \textbf{30.46} \\
CCS (Symmetric) & 0.5 & 0.5 & 5 & No & 29.01 \\
\midrule
CCS (Left-Skewed) & 0.7 & 0.3 & 1 & No & 21.78 \\
CCS (Right-Skewed) & 0.3 & 0.7 & 1 & No & 23.80 \\
CCS (Asymmetric Champion) & 0.1 & 0.5 & 1 & No & 25.97 \\
CCS (Asymmetric Low) & 0.1 & 0.1 & 1 & No & 22.96 \\
CCS (Asymmetric High) & 0.2 & 0.8 & 1 & No & 22.34 \\
CCS (Sweeter Spot) & 0.05 & 0.55 & 1 & No & 25.63 \\
CCS (Asymmetric + LH SI) & 0.1 & 0.5 & 1 & Yes & 16.27 \\
\midrule
LSCC (Symmetric K-Means) & K-Means & K-Means & 1 & No & 8.79 \\
LSCC (Symmetric K-Means) & K-Means & K-Means & 5 & No & 10.39 \\
\midrule
Hybrid (LSCC / CCS Asymmetric) & K-Means & 0.5 & 1 & No & 13.20 \\
Hybrid (LSCC / CCS Asymmetric) & K-Means & 0.5 & 2 & No & 17.58 \\
Hybrid (LSCC / CCS Asymmetric) & K-Means & 0.5 & 3 & No & 20.09 \\
Hybrid (LSCC / CCS Asymmetric) & K-Means & 0.5 & 4 & No & 18.72 \\
Hybrid (LSCC / CCS Asymmetric) & K-Means & 0.5 & 5 & No & \textbf{22.75} \\
Hybrid (LSCC / CCS, Seed 1001) & K-Means & 0.5 & 5 & No & 20.14 \\
\bottomrule
\end{tabular}%
}
\end{table}

\subsection{Split-CIFAR-100 Architecture Capacity and Parameter Scaling}\label{subsec:app_cifar_scaling}

To investigate the capacity limits and scaling robustness of 4MAS relative to B-IR, we evaluated both architectures across a range of total parameter capacities (from 14M to 140M parameters). The comparative results are summarized in Table~\ref{tab:app_cifar_scaling}.

\begin{table}[h]
\centering
\caption{Split-CIFAR-100 Class-IL performance across parameter capacity scales.}
\label{tab:app_cifar_scaling}
\resizebox{\linewidth}{!}{%
\begin{tabular}{llccc}
\toprule
\textbf{Model Size} & \textbf{Architecture} & \textbf{Total Parameters} & \textbf{Class-IL Acc (\%)} & \textbf{Forgetting (\%)} \\
\midrule
14M Scale & 4MAS (Ours, Scaled-down) & 14.1M & 13.16 & \textbf{25.8} \\
          & B-IR (Standard Baseline)* & 13.3M & \textbf{21.0} & 61.1 \\
\midrule
35M Scale & 4MAS (Ours, Scaled-down) & 34.5M & 18.41 & \textbf{25.8} \\
          & B-IR (Ours, Scaled-down) & 35.8M & \textbf{23.01} & 58.9 \\
\midrule
70M Scale & 4MAS (Ours, Champion) & 70.0M & $24.45 \pm 0.32$ & \best{$25.8 \pm 1.2$} \\
          & B-IR (Ours, Scaled) & 62.7M & \best{$26.27 \pm 1.37$} & $52.82 \pm 1.97$ \\
\midrule
140M Scale & 4MAS (Ours, Rank 1 Peak) & 130.8M & \best{$29.29 \pm 0.29$} & 25.8 \\
           & 4MAS (Ours, Peak v5)* & 147.8M & 29.01 & \textbf{23.4} \\
           & 4MAS (Ours, v10) & 130.8M & 27.18 & 73.7 \\
           & 4MAS (Ours, v9) & 130.8M & 27.13 & 51.0 \\
           & B-IR (Ours, Scaled) & 143.8M & 27.85 & 52.92 \\
\bottomrule
\end{tabular}%
}
\end{table}

As shown in Table~\ref{tab:app_cifar_scaling}, when properly initialized with pre-trained convolutional features (using the `--pre-convE` flag), the B-IR baseline scales robustly and monotonically with parameter capacity, rising from $21.0\%$ (13.3M scale) to $27.85\%$ (143.8M scale). Similarly, 4MAS scales monotonically and robustly across all parameter scales, rising from $13.16\%$ (14M scale) to $18.41\%$ (35M scale), $24.45 \pm 0.32\%$ (70M scale), and achieving a peak accuracy of \best{$29.29 \pm 0.29\%$} at 130.8M capacity (\texttt{fb=1.0, mb=20.0, bg=0.90}) under strict single-task pre-training constraints, outperforming the unilateral B-IR baseline as well as the 147.8M v5 configuration ($29.01\%$) which utilized multi-task pre-training. This breakthrough is achieved by resolving representation anchoring bottlenecks at scale via an expanded rehearsal buffer (\texttt{stmsize} = 4000). Importantly, 4MAS maintains significantly lower forgetting ($25.8 \pm 1.2\%$ at 70M scale and $23.4\%$ at 140M scale) than B-IR ($52.82 \pm 1.97\%$ and $52.92\%$), verifying that bilateral sleep consolidation consistently provides superior retention of past task knowledge across all capacity scales.

\subsection{VAE Hyperparameter Seed Replications and Task Trajectories}\label{subsec:app_vae_seeds}

\begin{figure}[h]
\centering
\includegraphics[width=0.95\linewidth]{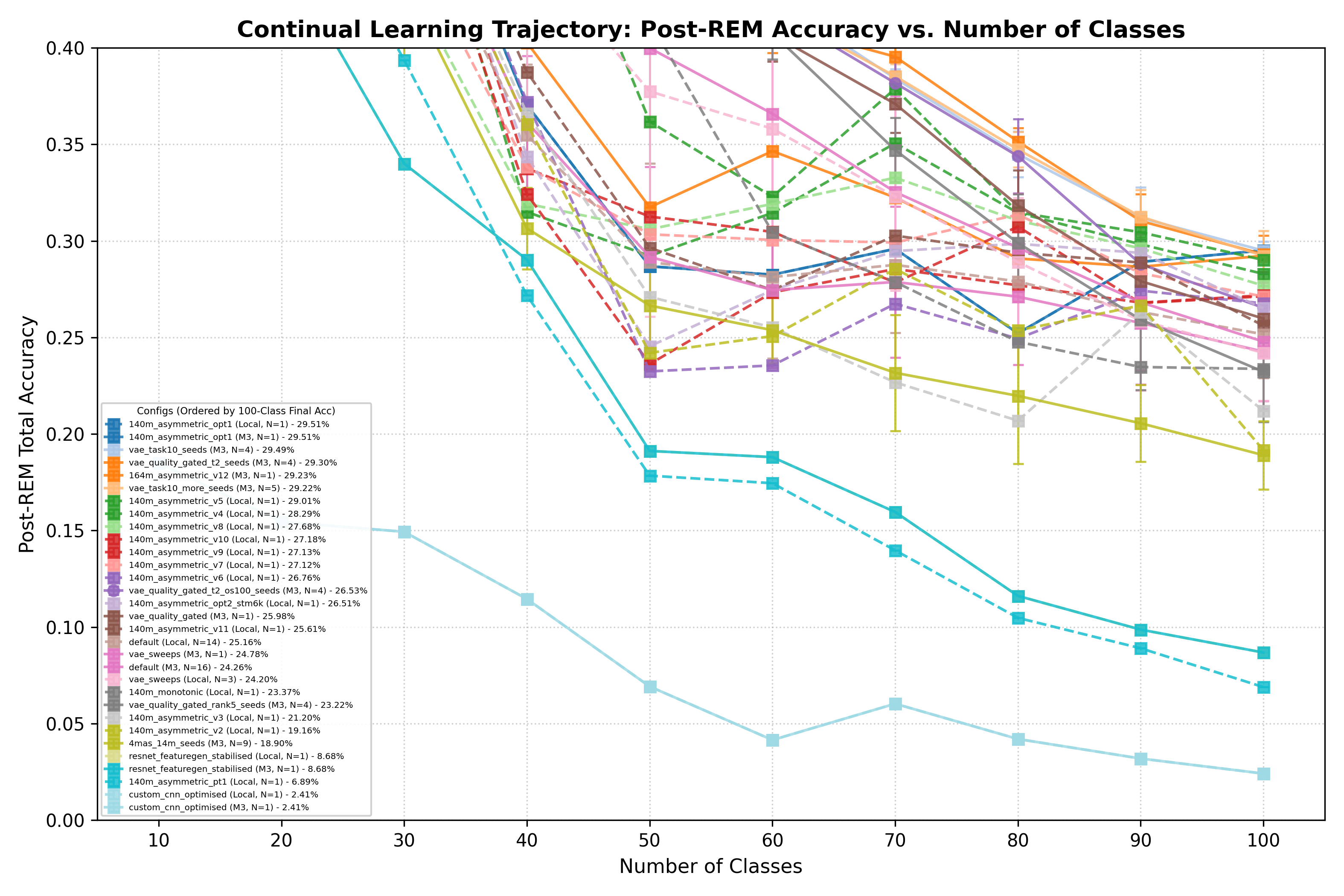}
\caption{Continual learning post-REM accuracy trajectories across class scale (10 to 100 classes) for top VAE configurations and Quality-Gated replay regimes, showing mean performance and standard deviation error bars across random seeds.}
\label{fig:vae_seeds_errorbars}
\end{figure}

To verify the statistical consistency and trajectory stability of top-performing VAE hyperparameter configurations and Quality-Gated rehearsal regimes, multiple random seed replications were evaluated across all 10 tasks on Split-CIFAR-100. Figure~\ref{fig:vae_seeds_errorbars} illustrates the post-REM joint accuracy trajectory as a function of class scale (10 to 100 classes) with standard deviation error bars.

\end{document}